\documentclass[jair,twoside,11pt,theapa]{article}
\usepackage{jair, theapa, rawfonts}

\usepackage{subcaption}
\usepackage{graphicx}
\usepackage{wrapfig}
\usepackage[dvipsnames]{xcolor}
\usepackage{amsmath}
\usepackage{mathtools}
\usepackage{tikz}
\usetikzlibrary{decorations.pathreplacing}
\usetikzlibrary{fit,shapes}
\usepackage{pifont}
\usepackage{booktabs}
\usepackage{bm}
\usepackage{mdframed}
\usepackage{url}
\usepackage{wrapfig}
\usepackage{amsfonts}
\usepackage{multirow}

\jairheading
\ShortHeadings{Improving Label Error Detection with Uncertainty Quantification}
{Jakubik, Vössing, Maskey, Wölfle, \& Satzger}
\firstpageno{1}

\begin{document}

\title{Improving Label Error Detection and Elimination with Uncertainty Quantification}
\author{\name Johannes Jakubik \email johannes.jakubik@kit.edu \\
      \addr Karlsruhe Institute of Technology\\
      Karlsruhe, Germany
      \AND
      \name Michael Vössing \email michael.voessing@kit.edu \\
      \addr Karlsruhe Institute of Technology\\
      Karlsruhe, Germany
      \AND
      \name Manil Maskey \email manil.maskey@nasa.gov \\
      \addr NASA IMPACT\\
      Marshall Space Flight Center, Huntsville, U.S.
      \AND
      \name Christopher Wölfle \email christopher.woelfle@kit.edu\\
      \addr Karlsruhe Institute of Technology\\
      Karlsruhe, Germany
      \AND
      \name Gerhard Satzger \email gerhard.satzger@kit.edu \\
     \addr Karlsruhe Institute of Technology\\
      Karlsruhe, Germany}


\maketitle

\begin{abstract}
\noindent Identifying and handling label errors can significantly enhance the accuracy of supervised machine learning models. 
Recent approaches for identifying label errors demonstrate that a low self-confidence of models with respect to a certain label represents a good indicator of an erroneous label.
However, latest work has built on softmax probabilities to measure self-confidence. 
In this paper, we argue that---as softmax probabilities do not reflect a model's predictive uncertainty accurately---label error detection requires more sophisticated measures of model uncertainty.  
Therefore, we develop a range of novel, model-agnostic algorithms for Uncertainty Quantification-Based Label Error Detection (UQ-LED), which combine the techniques of confident learning (CL), Monte Carlo Dropout (MCD), model uncertainty measures (e.g., entropy), and ensemble learning to enhance label error detection. 
We comprehensively evaluate our algorithms on four image classification benchmark datasets in two stages. 
In the first stage, we demonstrate that our UQ-LED algorithms outperform state-of-the-art confident learning in identifying label errors.  
In the second stage, we show that removing all identified errors from the training data based on our approach results in higher accuracies than training on all available labeled data. 
Importantly, besides our contributions to the detection of label errors, we particularly propose a novel approach to generate realistic, class-dependent label errors synthetically.
Overall, our study demonstrates that selectively cleaning datasets with UQ-LED algorithms leads to more accurate classifications than using larger, noisier datasets.
\end{abstract}

\section{Introduction}

In recent years, researchers and practitioners have mainly focused on improving the performance of machine learning models based on adjustments in the architectures and hyperparameters of the models. This so-called model-centric paradigm is increasingly complemented by the emerging data-centric paradigm, where the model---instead of the data---is considered static \cite{mazumder_dataperf_2022,hamid_model-centric_2022}. 
In contrast to improving the model, this approach aims at systematically enhancing the quality and quantity of data used \cite{aroyo_data_2021,wang_whose_2022,haakman_ai_2021,Sambasivan2021}. 
For supervised machine learning, an essential ingredient for strong performance is represented by high label quality \cite{han_survey_2020,wang_whose_2022}.
However, gathering data labels is not only costly but also prone to errors \cite{daniel_quality_2018,han_survey_2020,cordeiro_survey_2020}. 
Mislabeled data can then result in reduced performances of machine learning models, increased computational effort (e.g., driven by additionally required training data), and higher model complexity \cite{frenay_classification_2014}.

\begin{figure}[ht]%
\centering
\includegraphics[width=\textwidth]{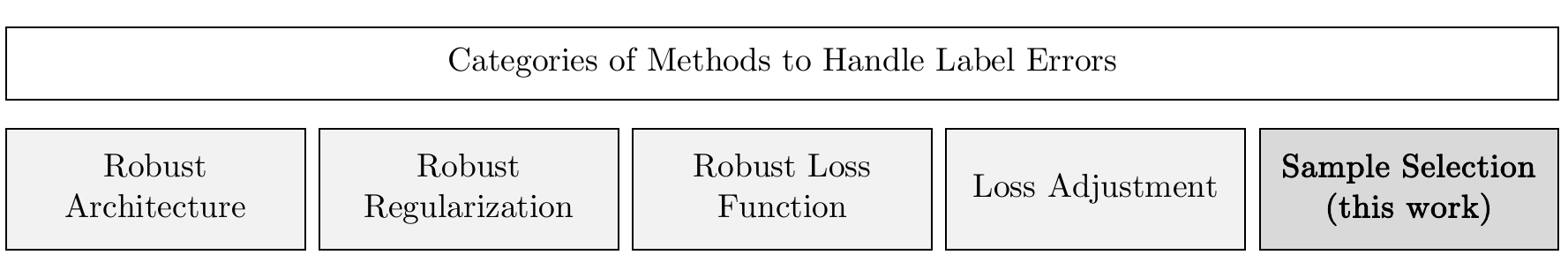}
\caption[Classification of state-of-the-art methods for handling label errors]{Classification of state-of-the-art methods for handling label errors. Adapted from \cite{song_learning_2022}.}
\label{fig:label_noise_methods}
\end{figure}

Recent work has demonstrated that label errors (often also called label noise) occur even in the most popular datasets \cite{northcutt_pervasive_2021}. For example, in MNIST images depicting the digit nine are labeled as the digit \textit{eight}, CIFAR-10 includes images of frogs, which are incorrectly labeled as cats, CIFAR-100 entails images depicting tigers, which are labeled as leopards, and in ImageNet images depicting meerkats are labeled as red pandas. 
Such incorrect labels affect the learning process of supervised machine learning models. 
Importantly, this especially applies to a vast majority of models pre-trained on standard datasets like ImageNet in the computer vision domain \cite{northcutt_pervasive_2021}. 
To mitigate the undesired downstream effects of mislabeled data, several techniques have emerged to account for label noise \cite{song_learning_2022}. 
Most of these approaches employ a model-centric perspective by trying to minimize the impact of label noise through specialized architectures \cite{xiao_learning_2015,goldberger_training_2017}, regularization techniques \cite{hendrycks_using_2019,lukasik_does_2020}, noise-robust loss functions \cite{ghosh_robust_2017,zhang_generalized_2018}, or loss adjustment \cite{patrini_making_2017,shu_meta-weight-net_2019}. 
In contrast, only a small set of works adopt a data-centric perspective and address the root cause of the problem---identifying and excluding erroneously labeled data. We summarize this in Figure~\ref{fig:label_noise_methods}. 

The approach of identifying and avoiding mislabeled data is called \textit{data cleansing}. 
One of the most prominent works in the field of data cleansing (both in research and practice), is represented by the confident learning (CL) technique \cite{northcutt_confident_2019}. 
{CL} encompasses a collection of model-agnostic algorithmic approaches that automatically identify and remove label errors in datasets. 
CL approaches utilize deep neural networks to compute the model's self-confidence for all labeled data, with the assumption that low self-confidence represents a good indicator for erroneous labels \cite{northcutt_confident_2019}. 
For the computation of self-confidence, CL utilizes out-of-sample softmax probabilities. 
However, in this work, we argue that softmax probabilities do not reflect a model's predictive uncertainty \cite{gal2016dropout} and, therefore, are not ideal candidates for self-confidence measures in label error detection. 
In contrast, we leverage work on accurately measuring the predictive uncertainty of ML models based on approximations of Bayesian neural networks and apply it in the context of label error detection. 
In this work, we specifically focus on Monte Carlo Dropout (MCD) for measuring predictive uncertainty due to the small required computational effort \cite{gal2016dropout}.

Our contributions are as follows: (1)~we improve CL by infusing more accurate estimates of the model uncertainty as self-confidence measures. 
Utilizing {MCD} probabilities allows us to account for the model's predictive uncertainty by including additional uncertainty measures, such as the entropy, in the label error detection process, thus improving the confidence of the algorithm's decision on the presence of a label error even under high uncertainty. 
We develop a range of algorithms for uncertainty quantification-based label error detection based on the different concepts and measurements of predictive uncertainty. 
(2)~We propose to use ensembles of algorithms (combining the novel algorithms with the state-of-the-art represented by the Prune-By-Noise-Rate Algorithm (CL-PBNR)) to leverage their individual strengths and, thereby, further improve the performance and generalizability of label error detection. 
(3)~We develop a novel approach of adding synthetic label errors to common {ML} benchmark datasets, which accounts for the distribution of classes and permutes the labels of most similar classes instead of random classes and, thereby, generates more realistic label errors compared to previous works \cite{northcutt_confident_2019}. 
(4)~We holistically evaluate our algorithms in a two-stage setup: First, we measure the accuracy of label error detection by comparing the data samples classified as label errors with our methods with the synthetically generated ground truth. 
Second, we exclude detected label errors from the training dataset and measure the effect on the final model accuracy with and without label errors to account for recent debates on the importance of well-curated datasets instead of big data.
Overall, our experiments demonstrate that incorporating predictive uncertainty based on approximations of Bayesian neural networks substantially improves the state-of-the-art in label error detection.

The remainder of this paper is structured as follows: 
First, we define label noise in the context of {ML}, outline different types of label errors, and highlight the resulting adverse consequences for the learning behavior. We then summarize CL as the current state-of-the-art method for identifying and excluding label errors and give a brief overview on approximations of Bayesian neural networks for uncertainty quantification. 
In Section~\ref{sec:uq}, we describe our approach of including uncertainty in label error detection and cover the algorithm development. 
In the subsequent section, we propose a novel, more realistic approach for the synthetic generation of label noise, which generates asymmetric and class-dependent label errors compared to recent approaches for randomly generated synthetic label noise. 
We then holistically evaluate the proposed algorithms in a two-stage evaluation setup (first on the test sets of common {ML} benchmark datasets, second by measuring the effects of removing label errors from the entire training data).
Finally, we discuss the implications and limitations of this work and provide an outlook on how future work can build on UQ-LED.

\section{Background}\label{sec:background}
Throughout this work, we focus on the classification of images as a very prominent and relevant task in the field of supervised {ML} \cite{bishop_pattern_2016}. We consider an ML model (also referred to as classifier), which has been trained on a specific dataset and, during inference, is utilized to classify unseen samples. 
We follow recent work and focus on classification tasks with multiple classes (i.e., no binary classification) using deep neural networks with a softmax layer as classifiers. 
The datasets considered in this work are usually labeled by humans, e.g., via crowdsourcing \cite{wang_crowdsourcing_2015,xiao_learning_2015,patrini_making_2017,cheng_learning_2017,daniel_quality_2018}. 

\subsection{Notation and Types of Label Errors}
The notation of this work follows to large degrees the notation of \cite{northcutt_confident_2019}.
Let $\mathcal{K}:=\{1,2,\dots, c\}$ denote the set of $c$ unique class labels  and $\bm{X}:= (\bm{x},\Tilde{y})^n \in (\mathbb{R}^{d}, \mathcal{K} )^{n}$ the training dataset consisting of $n$ samples. 
Each sample $\bm{x} \in \bm{X}$ has an observed noisy label $\Tilde{y} \in \mathcal{K}$, which associates the sample to one of the $c$ unique classes. 
While the observed label $\Tilde{y}$ should usually correspond to the true label $y^*$ \cite{frenay_classification_2014}, it is possible that before observing $\Tilde{y}$ a stochastic noise process \cite{angluin_learning_1988} maps $y^{\ast} \rightarrow \Tilde{y}$, e.g., a human mislabels a picture of a crow as a raven. 
During this process, every sample $\bm{x}$ with a true label $j \in \mathcal{K}$ may be independently mislabeled as class $i \in \mathcal{K}$ with a probability of $p(\Tilde{y}=i \mid y^{\ast}=j, \bm{x})$ \cite{song_learning_2022}. 
We will refer to this probability as the flipping probability.
Probabilities that have been predicted by a model $\bm{\theta}$ are denoted as $\hat{p}$. 
The $n \times m$ matrix of predicted probabilities is $\bm{\hat{P}}_{z,i}=\hat{p}(\Tilde{y}=i;\bm{x}_z,\bm{\theta})$, where $\bm{x}_z$ is the $z^{th}$ sample in the training dataset $\bm{X}$. 
The corresponding observed noisy label of sample $\bm{x}_z$ is $\Tilde{y}_z$. 
Sometimes we abbreviate $\hat{p}(\Tilde{y}=i;\bm{x}_z,\bm{\theta})$ with $\hat{p}_{\bm{x}_z,\Tilde{y}=i}$ or just $\hat{p}_{\bm{x},\Tilde{y}=i}$.
Lastly, let $\bm{\hat{p}}_{\bm{x}}:=\{ \hat{p}_{\bm{x},\Tilde{y}=i}, \forall i \in \mathcal{K} \}$ denote the vector of all softmax probabilities of sample $\bm{x} \in \bm{X}$ and let $\bm{X}_{\Tilde{y}=i}$ denote the subset of samples in the dataset belonging to class $i \in \mathcal{K}$.

\begin{figure}[ht]%
\centering
\includegraphics[width=0.8\textwidth]{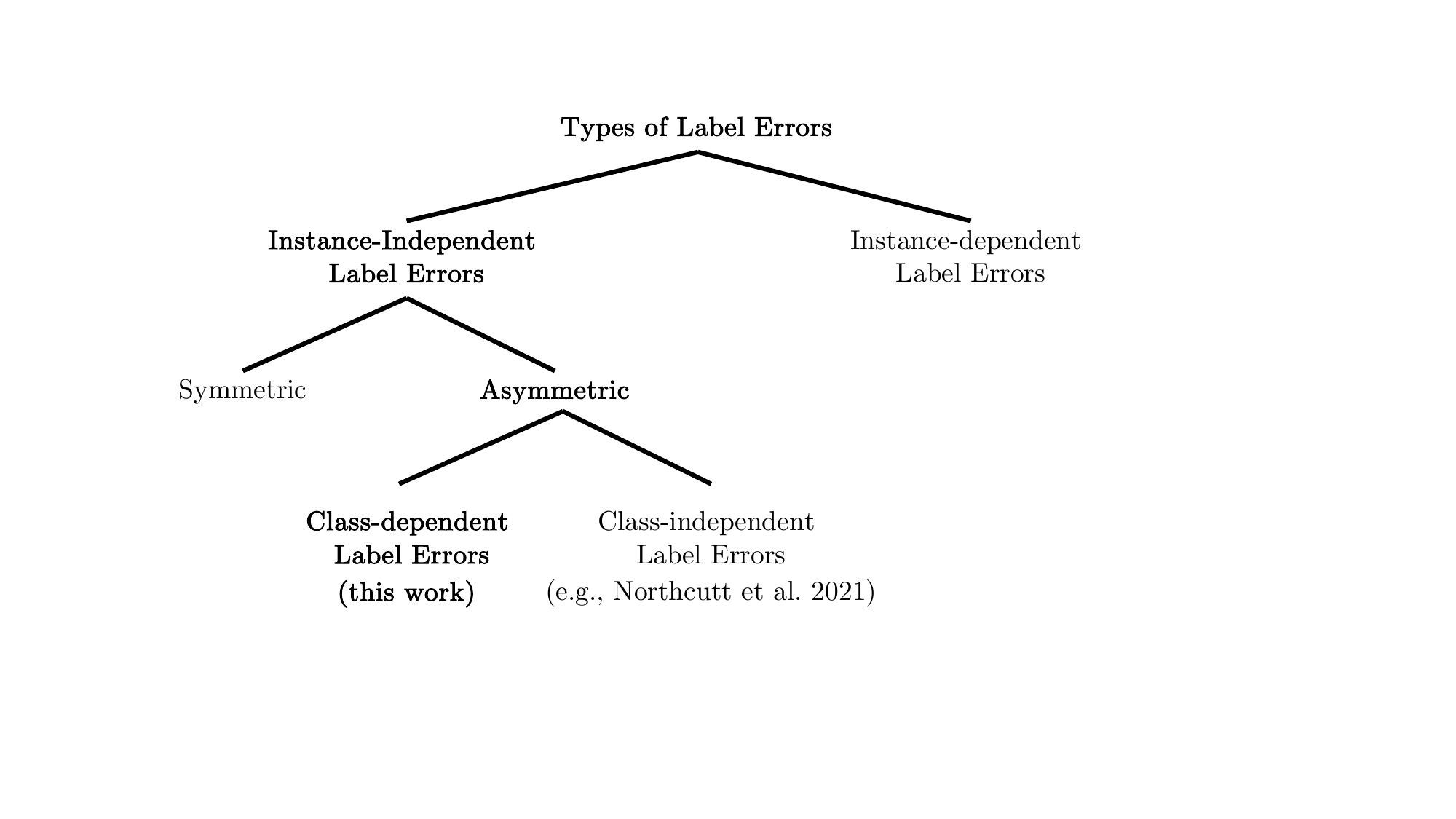}
\caption[Types of label errors]{Types of label errors.}
\label{fig:label-error-categories}
\end{figure}

Following \cite{cordeiro_survey_2020}, we define label errors (or label noise) as samples whose observed labels do not represent the true class \cite{cordeiro_survey_2020}. 
We note that there exist different types of label errors, which we cover in the following and summarize in Figure~\ref{fig:label-error-categories}. 

In the literature, different types of label errors have been considered \cite{angluin_learning_1988,frenay_classification_2014,patrini_making_2017,algan_label_2020,cordeiro_survey_2020,cheng_learning_2020,song_learning_2022}.  
Following the taxonomy of \cite{song_learning_2022}, the following label error types can be denoted: 1) instance-independent and 2) instance-dependent label errors. 
In the following, we will examine those different types and describe ways to model the respective noises.

\textbf{1) Instance--Independent Label Errors:} In this category, it is assumed that the noise process is independent of data features \cite{natarajan_learning_2013,xia_are_2019,han_survey_2020,song_learning_2022}.
In general, instance-independent label errors are characterized by a noise transition matrix $\bm{T} \in [0,1]^{c \times c}$, where $\forall \, i,j \in \mathcal{K}$
\begin{equation}
   \bm{T}_{ij}:=p\left( \Tilde{y}=j \mid y^*=i \right)
\end{equation}
denotes the flipping probability of the true label $i$ being mislabeled as class $j$ \cite{song_learning_2022}. 
This flipping probability is instance-independent meaning that $p\left( \Tilde{y}=j \mid y^*=i,\bm{x} \right)=p\left( \Tilde{y}=j \mid y^*=i \right)$ \cite{han_survey_2020}. 
There are two specific forms of instance-independent label errors: symmetric noise and asymmetric noise.

\textit{Symmetric noise}, also known as random noise or uniform noise, is a noise process where each label has an equal chance of flipping to another class with a fixed probability, independent of the true label $y^{*}$ \cite{cordeiro_survey_2020,cheng_learning_2017,song_learning_2022}. 
Therefore, all entries in the noise transition matrix are evenly distributed, except for the diagonal ones \cite{algan_label_2020}. Depending on the noise rate (ratio of noisy samples) $\tau \in [0,1]$, the noise transition matrix can simply be modeled $\forall \, i,j \in \mathcal{K}$ as 
\begin{align}
&\bm{T}_{ij}= 
\begin{cases}
    1-\tau,                       & \text{if } i=j, \\
    \frac{\tau}{c-1},             & \text{otherwise.}
\end{cases}
\end{align}

\textit{Asymmetric noise}, also known as label-dependent or class-conditional noise, is a noise process where the flipping probability $p(\Tilde{y}=j \mid y^{\ast}=i)$ depends on the true label $y^{*}$ and thus is not the same for every class \cite{patrini_making_2017,natarajan_learning_2013,algan_label_2020,song_learning_2022}.
This type of noise is more comparable to label errors that occur in real life when classes are too similar for a human labeler regardless of the instance \cite{patrini_making_2017,cordeiro_survey_2020}. 
For example, an image of a donkey is more likely to be mislabeled as a horse than as a car. 
When modeling an asymmetric transition matrix for the noise rate $\tau \in [0,1]$, it must fulfill the following criteria, $\forall_{i=j} \bm{T}_{ij}=1-\tau \land 	\exists_{i \neq j, i \neq k, j \neq k} \bm{T}_{ij} > \bm{T}_{ik}$, with $i,j,k \in \mathcal{K}$, meaning that a true label has a higher chance to be mislabeled into a specific class \cite{song_learning_2022}. 
\cite{northcutt_confident_2019} propose a method for modeling synthetic asymmetric label errors that randomly generates transition matrices with varying degrees of sparsity (fraction of zeros in the off-diagonals of the transition matrix). 
A high degree of sparsity reflects asymmetric label errors \cite{northcutt_confident_2019}. 
However, their approach does not consider class similarities.
To model an asymmetric noise process that considers class similarities, \cite{algan_label_2020} propose to train a deep neural network ({DNN}) on the training set and then calculate its confusion matrix on the test set. The confusion matrix is then used as the noise transition matrix. \cite{gare_exploiting_2021} propose a similar approach to synthetically generate class-conditional label errors, which also considers class similarities. 
We will use their approach as the foundation of our own synthetic noise generator.

\textbf{2) Instance--Dependent Label Errors:}
Unlike instance-independent label errors, not all instances of a particular class have the same flipping probability. 
Instance-dependent or feature-dependent label errors is a more general noise process where the flipping probability $p(\Tilde{y}=i \mid y^{\ast}=j, \bm{x})$ depends on the true class label and the sample's features \cite{cheng_learning_2017,patrini_making_2017}. 
For example, a sample that has less information or is of lower quality is more likely to be mislabeled by a human \cite{cheng_learning_2017}. Instance-dependent noise is harder to model than the aforementioned noise types since similarities between all samples need to be calculated \cite{algan_label_2020,song_learning_2022}. 
For the complete notation of instance-dependent label errors, we refer to the works of \cite{xiao_learning_2015} and \cite{cheng_learning_2020} since this type of noise is not in the scope of this work. 

We will describe the differences between class-dependent and class-independent label errors throughout Section \ref{sec:label-noise}, where we introduce the generation of label noise for this work.

\subsection{Confident Learning}
\label{subsec:confident_learning}

Confident Learning (CL) represents the state-of-the-art in sample selection-based label error detection (see Figure~\ref{fig:label_noise_methods}).
In general, {CL} is a model-agnostic approach for characterizing, identifying, and learning with label errors, which makes use of the out-of-sample predicted probabilities to estimate the joint distribution between noisy and correct labels \cite{northcutt_confident_2019}. 
The joint distribution is then consulted to identify potential label errors.
These potential label errors are then excluded from the training data. 

Due to the relevance of confident learning for the remainder of this work, we describe the approach by \cite{northcutt_confident_2019} in greater detail in the following:
The general assumption of CL is that label noise is class-conditional, meaning that $p(\Tilde{y} \mid y^{\ast}, \bm{x}) = p(\Tilde{y} \mid y^{\ast})$. To characterize this class-conditional noise, the latent (unknown) prior distribution of correct labels $p(\Tilde{y} \mid y^{\ast})$ and $p(y^{\ast})$ needs to be estimated. 
Instead of estimating $p(\Tilde{y} \mid y^{\ast})$ and $p(y^{\ast})$ separately, CL estimates both directly by estimating the joint distribution of the label errors $p( \Tilde{y}, y^{\ast} )$\footnote{Product Rule: $p(a,b)=p(a \mid b) \cdot p(b)$} as matrix $\bm{Q}_{\Tilde{y}, y^{\ast} }$.

CL requires two inputs for detecting label errors. 
That is, the out-of-sample predicted softmax probabilities $\bm{\hat{P}}_{z,i}$ and the given labels of all samples $\Tilde{y}_z$.
The inputs are associated with each other via index $z$, $\forall \, x_z \in \bm{X}$. 
The out-of-sample predicted (softmax) probabilities $\bm{\hat{P}}_{z,i}$ can be obtained, for example, via cross-validation. 
Note that, in this work, we will change the input of CL and employ uncertainty measures instead of softmax probabilities as the input to calculate self-confidence.

{CL} then follows two major steps: First, the confident joint $\bm{C}_{\Tilde{y},y^{\ast}}$ is determined. This probability estimates $\bm{X}_{\Tilde{y}=i,y^{\ast}=j}$, the set of samples $\bm{x}$ labeled $\Tilde{y}=i$ with the probability $\hat{p}(\Tilde{y}=j;\bm{x},\bm{\theta})$ being high enough to be confidently counted as class $y^{\ast}=j$.
In other words, it is the set of samples that likely belong to class $j$ although their given noisy label is $\Tilde{y}=i$.
This decision is based on a class-specific threshold $t_j$, which represents the expected average self-confidence $\forall k \in \mathcal{K}$.
\cite{northcutt_confident_2019} define self-confidence as the predicted probability that a sample $\bm{x}$ belongs to its given label $\Tilde{y}$, which can be expressed as $\hat{p}(\Tilde{y}=i;\bm{x}\in X_{\Tilde{y}=i}, \bm{\theta})$. 
Low self-confidence is then considered an indicator of a label error. 
After the confident joint has been calculated for every sample $\bm{x} \in \bm{X}$ the joint distribution matrix $\bm{Q}_{\Tilde{y}, y^{\ast}}$ can be estimated (see \cite{northcutt_confident_2019} for details).
Second, after the confident joint $\bm{C}_{\Tilde{y},y^{\ast}}$ and the joint distribution $\bm{Q}_{\Tilde{y}, y^{\ast}}$ have been estimated, various rank and prune algorithms can be utilized to determine which samples should be excluded from the data. 
Recent work has demonstrated that the so-called \textit{Prune by Noise Rate (PBNR)} approach achieves particularly high accuracy in identifying label errors \cite{northcutt_confident_2019}. 
PBNR essentially measures the margin of how certain the model is that the given label $\Tilde{y}=i$ is false, assuming that the true label is $y^{*}=j$. 
For each off-diagonal entry in $\bm{\hat{Q}}_{\Tilde{y}=i, y^{\ast}=j}$, where $i \neq j$, remove the $n \cdot \bm{\hat{Q}}_{\Tilde{y}=i, y^{\ast}=j }$ samples with the largest margin $\hat{p}_{\bm{x},y^{*}=j} - \hat{p}_{\bm{x},\Tilde{y}=i}$.

\subsection{Uncertainty in Deep Learning}
\label{subsec:mc_dropout}
Deep neural networks ({DNN}) do not capture the model's predictive uncertainty during a classification task. 
In contrast, DNNs calculate predictive probabilities obtained as output during inference by a {DNN} with a softmax function \cite{gal2016dropout}. 
These softmax probabilities $\hat{p}_{\bm{x},\Tilde{y}=k}$ indicate how likely a sample $\bm{x}$ belongs to one of the $k \in \mathcal{K}$ classes, by approximating the relative probabilities between the different classes \cite{kendall_bayesian_2015,gal2016dropout}. 
Contrary to common belief, those probabilities do not reflect the model's predictive uncertainty \cite{gal2016dropout,hendrycks_baseline_2016,northcutt_confident_2019}, meaning that even with a high softmax output, a model can be uncertain \cite{gal2016dropout}. 

One approach to model predictive uncertainty is to use Bayesian neural networks (BNN). 
The Bayesian probability theory provides well-established methods like BNNs and Gaussian processes for obtaining model uncertainty in addition to the predictive probabilities. 
BNNs are stochastic neural networks that place distributions over the model parameters that express how likely the different parameter values are \cite{neal_bayesian_1996,jospin_hands-bayesian_2022}. This approach differs from conventional deep neural networks, which assign each parameter a single value obtained by minimizing a loss function \cite{jospin_hands-bayesian_2022}.
So unlike deep neural networks, BNNs provide predictive distributions as outputs, which can be used to obtain model uncertainty \cite{alarab_illustrative_2021,gal2016dropout}.
However, BNNs are computationally expensive. 
To mitigate the computational costs of BNNs, \cite{gal2016dropout} have proposed lightweight approximations of BNNs based on the so-called Monte Carlo Dropout (MCD). 

Dropout is typically applied as a regularization technique at training time to reduce overfitting \cite{srivastava_dropout_2014}. 
During inference, dropout is usually not employed \cite{srivastava_dropout_2014}. 
However, recent work has demonstrated that drawing Monte Carlo samples from a model based on enabled dropout inference approximates {BNN}. 
During {MCD} inference, a probability distribution is generated by drawing multiple predictive samples from the model over various stochastic forward passes. During each forward pass, different neurons are switched off, leading to slightly different softmax outputs \cite{gal2016dropout}.
Formally, let $\hat{p}_{\bm{x},\Tilde{y}=k,f}=\hat{p}_f(\Tilde{y}=k;\bm{x},\bm{\theta})$ denote the softmax output of model $\bm{\theta}$ of forward pass $f$ for class $k \in \mathcal{K}$. To get the final predictive output (per-class {MCD} probabilities), the mean over the outcomes of all $F$ forward passes is calculated \cite{gal2016dropout} as 
\begin{equation}
    \hat{p}^{\textbf{MCD}_{\bm{x},\Tilde{y}=k}}=\frac{1}{F} \sum_{f=1}^{F}\hat{p}_{\bm{x},\Tilde{y}=k,f} \quad \forall k \in \mathcal{K},
\end{equation}
while the final classification $\hat{y}$ is obtained as
\begin{equation}
    \hat{y}=\underset{k \in \mathcal{K}}{\mathrm{arg\,max}}\left( \hat{p}^{\textbf{MCD}_{\bm{x},\Tilde{y}=k}} \right).
\end{equation}

Since {MCD} inference provides a probability distribution that reflects an estimation of the model's uncertainty \cite{alarab_illustrative_2021},
this probability distribution can be used to calculate uncertainty measures, such as the entropy \cite{gal2016dropout}, which is defined as 
\begin{equation}
    H(\bm{\hat{p}}_{\bm{x}}^{\text{MCD}})=-\sum_{k=1}^{K} \hat{p}^{\text{MCD}}_{\bm{x},\Tilde{y}=k} \cdot \log \left( \hat{p}^{\text{MCD}}_{\bm{x},\Tilde{y}=k} \right).
\end{equation}
For a uniform distribution of probabilities across classes, the entropy reaches its maximum. In contrast, if one class $i \in \mathcal{K}$ has a probability of $\hat{p}^{\textbf{MCD}_{\bm{x},\Tilde{y}=i}}=1$, while the other classes $j \in \mathcal{K} \land j \neq i$ have a probability of $\hat{p}^{\textbf{MCD}_{\bm{x},\Tilde{y}}=j}=0$, the entropy is minimal \cite{treiss_uncertainty-based_2021}.

\section{Infusing Model Uncertainty in Label Error Detection}\label{sec:uq}

In this work, we develop novel, model-agnostic, uncertainty-based algorithms for label error detection. For this, we follow \cite{northcutt_confident_2019} and train an ML model on labeled data, including erroneous labels. We then utilize our developed algorithms to identify the label errors and measure both the label detection accuracy and the effect of excluding erroneous labeled data from training on the final model accuracy.

For our algorithms, we build on the previously introduced CL-PBNR algorithm by \cite{northcutt_confident_2019}. 
However, instead of using out-of-sample predicted softmax probabilities, we leverage {MCD} probabilities, which allow us to employ uncertainty measures based on the underlying approximation of Bayesian neural networks resulting from multiple stochastic forward passes.
On top of adding MCD probabilities, we then incorporate uncertainty measures, e.g., the entropy $H$ or margin-based techniques, to improve the detection performance based on higher precision in assessing the confidence. 
In the following, we introduce the five developed algorithms for UQ-LED.

\textbf{CL-MCD}
In order to measure the effect of using {MCD} as an approximation of Bayesian neural networks in label error detection, we replace the softmax probabilities with {MCD} probabilities as input to the CL-PBNR algorithm by \cite{northcutt_confident_2019}. 

\textbf{CL-MCD + Entropy}
On top of MCD probabilities, the second algorithm leverages the entropy $H$ of the MCD probabilities as input to the CL-PBNR algorithm. Following the recommendations for future research of \cite{northcutt_confident_2019} to try other thresholds, we use the entropy as an additional per-class threshold during the computation of the confident joint $\bm{C}_{\Tilde{y},y^{\ast}}$. 
The entropy threshold $te_j$ for each class $j \in \mathcal{K}$ is defined as the average entropy of all samples where the given label is $j$:
\begin{equation}
    te_j=\frac{1}{\lvert \bm{X}_{\Tilde{y}=j} \rvert} \sum_{\bm{x} \in \bm{X}_{\Tilde{y}=j}} H\left( \bm{\hat{p}}^{\text{MCD}}_{\bm{x}} \right) \, .
\end{equation}
Since the entropy converges to zero with increasing model confidence about the prediction, the threshold is fulfilled if 
\begin{equation}
    H\left( \bm{\hat{p}}^{\text{MCD}}_{\bm{x} \in \bm{X}_{\Tilde{y}=j}} \right) \leq te_j \, .
\end{equation}

A sample is now counted as confident prediction inside the confident joint if the default threshold $t_j$ (average self-confidence for each class) defined in Section \ref{subsec:confident_learning}, and the new entropy threshold $te_j$ is fulfilled. 
In other words, a sample is now only counted if the model is confident that the given label is wrong and if the model is certain with its overall prediction for the given sample.
In the following, we will abbreviate this algorithm as CL-MCD-E.

\begin{figure}[h]%
\centering
\includegraphics[width=0.95\textwidth]{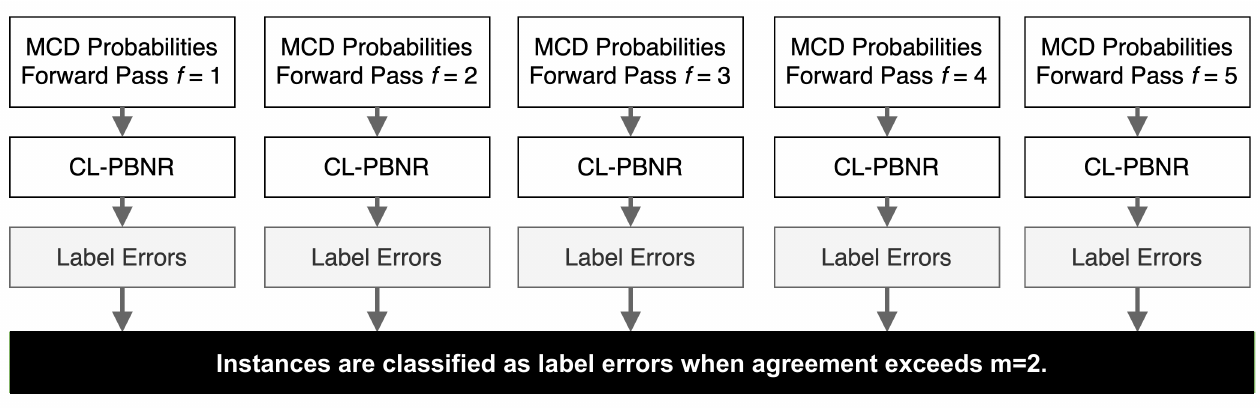}
\caption[Visualization of an ensemble based on the combination of confident learning and Monte Carlo Dropout]{Overview of the CL-MCD-Ensemble.}
\label{fig:alg-4}
\end{figure}

\textbf{CL-MCD-Ensemble}
When performing {MCD} inference, the results of $F$ forward passes are combined into one single probability (see \ref{subsec:mc_dropout}). 
Each forward pass $f$ leads to a (slightly) different probability. 
By using the principles of ensemble learning, we feed each of the $F$ predictions into the CL-PBNR algorithm and thus create a homogenous ensemble. 
Each of the $F$ ensemble members receives the {MCD} probabilities from a different forward pass $f$ as input and then returns a list of possible label error candidates. 
Based on the principle of majority vote, a sample's label is considered corrupted if more than 50\% of the ensemble members marked the sample as a potential label error. 
Let $m$ denote the required minimum number of ensemble members that must have identified a sample as erroneous in order to actually mark the sample as a label error. 
So if the number of forward passes is $F=5$, then $m$ must equal three in order to satisfy the majority vote condition.   
An overview of the CL-MCD-Ensemble can be found in Figure~\ref{fig:alg-4}.

\begin{figure}[h]%
\centering
\includegraphics[width=0.9\textwidth]{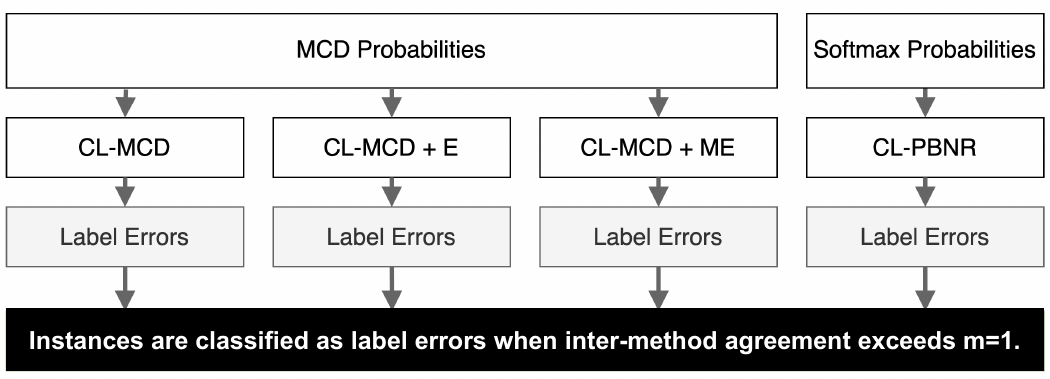}
\caption[Overview of the ensemble of algorithms]{Overview of the Algorithm Ensemble.}
\label{fig:alg-5}
\end{figure}

\textbf{Algorithm Ensemble}
Having a homogenous ensemble at hand, we also want to explore a heterogenous ensemble consisting of several different algorithms. 
Therefore, we use our new algorithms 1-3 plus the CL-PBNR algorithm, resulting in an ensemble with four members. 
We included the CL-PBNR method as we noticed during the development that it has a very good recall compared to our proposed algorithms, which excel in their precision.
The CL-PBNR algorithm receives softmax probabilities, and our other uncertainty-based algorithms receive {MCD} probabilities and, if required, the entropy as input. 
For this ensemble, we also apply the principle of majority vote. 
So, a sample's label is considered corrupted if more than 50\% of the four ensemble members ($m=3$) marked the sample as a potential label error.  
As this ensemble only consists of four members, we will additionally examine the label error detection performance for $m=2$.
A visual overview of the ensemble can be found in \ref{fig:alg-5}. 
In the following, we will abbreviate this ensemble with CL-MCD-Ens.

\section{Generating Realistic Label Noise}\label{sec:label-noise}

For the evaluation of the label error detection, synthetic label errors need to be added to the datasets as, to our knowledge, there is no ground-truth dataset for label error detection.
We follow the work on confident learning by considering asymmetric label noise, where some label errors are more likely than others. 
However, we deviate from \cite{northcutt_confident_2019} by considering class similarities in asymmetric label noise---e.g., a cat is more likely to be confused with a dog than a car. 
We follow the literature on class-dependent label noise and train a model on the training set, and then use the predictions on the test set to generate similarity scores \cite{gare_exploiting_2021}. 
These similarity scores indicate how similar one class is to another. 
Following \cite{gare_exploiting_2021}, we define the similarity score between class $k$ and class $l$, for all $\forall k,l \land k \neq l$ as
\begin{equation}
    S_{k,l} = \frac{1}{\lvert \bm{X}_{\Tilde{y}=k} \rvert} \sum_{\bm{x} \in \bm{X}_{\Tilde{y}=k}} \hat{p}(\Tilde{y}=l;\bm{x},\bm{\theta}),
\end{equation}
i.e., the average of all probabilities of class $l$ where the given label $\Tilde{y}$ equals $k$.
The set of similarity scores for class $k$ is defined as
\begin{equation}
    \bm{S}_k[l] := \{ S_{k,l}, \, \, \forall  \, l \in \mathcal{K} \land l \neq k\}
\end{equation}

To better understand the noise generation process, we provide an example by adding label noise to the CIFAR-10 dataset based on the \textit{airplane} class. 
In Table~\ref{tab:similarity-score-set}, we report the similarity scores on images from the test data of class $k$=\textit{airplane}.

\begin{table}[h]
\caption[Example of class similarities represented by softmax scores of deep learning models]{Example of class similarities represented by softmax scores of deep learning models for class \textit{k=airplane} of CIFAR-10 dataset.}
\label{tab:similarity-score-set}
\resizebox{\textwidth}{!}{%
\begin{tabular}{|l|c|c|c|c|c|c|c|c|c|}
\hline
$l$                             & \textbf{automobile} & \textbf{bird} & \textbf{cat} & \textbf{deer} & \textbf{dog} & \textbf{frog} & \textbf{horse} & \textbf{ship} & \textbf{truck} \\ \hline
$S_{\text{airplane},l}$   & 0.004               & 0.019          & 0.007         & 0.002          & 0.000         & 0.001          & 0.002           & 0.027          & 0.010           \\ \hline
\end{tabular}
}
\end{table}

After the similarity scores have been computed, a threshold
\begin{equation}
    ts_{k}= \overline{\bm{S}_k} + \sigma(\bm{S}_k) 
\end{equation}
is calculated, and the indexes $l$ of all similarity score set entries $\bm{S}_k[l]$ that are higher than the threshold $ts_k$ are grouped in the similarity group $\mathcal{G}_k$, which is defined as 
\begin{equation}
    \mathcal{G}_k :=\{ \, l : \bm{S}_k[l] \geq ts_k,  \, \forall  \, l \in \mathcal{K} \land l \neq k \}.
\end{equation}

For the CIFAR-10 example, the threshold $ts_{\text{airplane}}=0.016$ and since only $S_{\text{airplane},\text{bird}}$ and $S_{\text{airplane},\text{ship}}$ are greater than the threshold the similarity group for class $k=\text{\textit{airplane}}$ equals 
\begin{equation}
    \mathcal{G}_{\text{airplane}}:=\{\text{\textit{bird}, \textit{ship}}\}.
\end{equation}

After the similarity group $\mathcal{G}_k$ is available, the synthetic flipping probabilities $\bm{FP}_k$ for class $k$ are calculated as 
\begin{align}
&\bm{FP}_k[i]:= 
\begin{cases}
    \mathrm{softmax} \left(\bm{S}_k[l] \right),   & \text{if } l \in \mathcal{G}_k \\
    0,                          & \text{otherwise,}
\end{cases}
\end{align}
with $i \in \mathcal{K}$. 
All the aforementioned steps are repeated for all classes $k \in \mathcal{K}$.

For the CIFAR-10 example, the synthetic flipping probabilities $\bm{FP}_{\text{airplane}}$ are reported in Table~\ref{tab:flipping-probs}. 

\begin{table}[h]
\caption[Class-dependent flipping probabilities for the generation of synthetic label errors]{Class-dependent flipping probabilities for the generation of synthetic label errors for all instances with given label $\Tilde{y}=k=$\textit{airplane}.}
\label{tab:flipping-probs}
\resizebox{\textwidth}{!}{%
\begin{tabular}{|l|c|c|c|c|c|c|c|c|c|c|}
\hline
$i$  & \textbf{airplane} & \textbf{automobile} & \textbf{bird} & \textbf{cat} & \textbf{deer} & \textbf{dog} & \textbf{frog} & \textbf{horse} & \textbf{ship} & \textbf{truck} \\ \hline
$\bm{FP}_{\text{airplane}}$     & 0.0\%            & 0.0\%              & 49.8\%        & 0.0\%            & 0.0\%             & 0.0\%            & 0.0\%             & 0.0\%              & 50.2\%        & 0.0\%              \\ \hline
\end{tabular}
}
\end{table}
Finally, to generate label noise for different noise rates $\tau$, $\tau \cdot \lvert \bm{X} \rvert$ instances are randomly selected whose label will be flipped. 
The label of each selected instance $\bm{x} \in \bm{X}$ with given label $\Tilde{y}=k$ is flipped based on the previously calculated probabilities $\bm{FP}_k$. 

In Table~\ref{tab:cifar10-noise-transition}, the noise transition matrix generated by our asymmetric noise generator for the CIFAR-10 dataset with a noise rate of $\tau=0.2$ is shown.
For our CIFAR-10 example, an instance with given label $\Tilde{y}=k= \text{\textit{airplane}}$  is misclassified as \textit{bird} or \textit{ship} with the same probability. 
The misclassification as \textit{bird} seems plausible. 
The misclassification as a \textit{ship} may come from the fact that the CIFAR-10 dataset also contains images of seaplanes that can easily be mistaken for a ship.

\begin{table}[h]
\caption[Transition matrix $\bm{T}_{i,j}$ for synthetic label errors]{Transition matrix $\bm{T}_{i,j}$ for synthetic label errors for the CIFAR-10 dataset with a noise rate of $\tau=0.2$, e.g., instances with given label $i=$\textit{automobile} are mislabeled with a probability of 20.1\% as $j=$\textit{truck}.}
\label{tab:cifar10-noise-transition}
\resizebox{\textwidth}{!}{%
\begin{tabular}{|l|r|r|r|r|r|r|r|r|r|r|}
\hline
$\bm{T_{ij}}$       & \multicolumn{1}{c|}{\textbf{\begin{tabular}[c]{@{}c@{}}air-\\ plane\end{tabular}}} & \multicolumn{1}{c|}{\textbf{\begin{tabular}[c]{@{}c@{}}auto-\\ mobile\end{tabular}}} & \multicolumn{1}{c|}{\textbf{bird}} & \multicolumn{1}{c|}{\textbf{cat}} & \multicolumn{1}{c|}{\textbf{deer}} & \multicolumn{1}{c|}{\textbf{dog}} & \multicolumn{1}{c|}{\textbf{frog}} & \multicolumn{1}{c|}{\textbf{horse}} & \multicolumn{1}{c|}{\textbf{ship}} & \multicolumn{1}{c|}{\textbf{truck}} \\ \hline
\textbf{airplane}   & 80.3\%                                                        & 0.0\%                                                           & 9.9\%         & 0.0\%        & 0.0\%         & 0.0\%        & 0.0\%         & 0.0\%          & 9.9\%         & 0.0\%          \\ \hline
\textbf{automobile} & 0.0\%                                                         & 79.9\%                                                          & 0.0\%         & 0.0\%        & 0.0\%         & 0.0\%        & 0.0\%         & 0.0\%          & 0.0\%         & 20.1\%         \\ \hline
\textbf{bird}       & 0.0\%                                                         & 0.0\%                                                           & 80.4\%        & 10.0\%       & 9.7\%         & 0.0\%        & 0.0\%         & 0.0\%          & 0.0\%         & 0.0\%          \\ \hline
\textbf{cat}        & 0.0\%                                                         & 0.0\%                                                           & 0.0\%         & 78.9\%       & 0.0\%         & 21.1\%       & 0.0\%         & 0.0\%          & 0.0\%         & 0.0\%          \\ \hline
\textbf{deer}       & 0.0\%                                                         & 0.0\%                                                           & 0.0\%         & 19.3\%       & 80.7\%        & 0.0\%        & 0.0\%         & 0.0\%          & 0.0\%         & 0.0\%          \\ \hline
\textbf{dog}        & 0.0\%                                                         & 0.0\%                                                           & 0.0\%         & 19.0\%       & 0.0\%         & 81.0\%       & 0.0\%         & 0.0\%          & 0.0\%         & 0.0\%          \\ \hline
\textbf{frog}       & 0.0\%                                                         & 0.0\%                                                           & 0.0\%         & 20.2\%       & 0.0\%         & 0.0\%        & 79.8\%        & 0.0\%          & 0.0\%         & 0.0\%          \\ \hline
\textbf{horse}      & 0.0\%                                                         & 0.0\%                                                           & 0.0\%         & 6.7\%        & 7.0\%         & 6.6\%        & 0.0\%         & 79.7\%         & 0.0\%         & 0.0\%          \\ \hline
\textbf{ship}       & 20.5\%                                                        & 0.0\%                                                           & 0.0\%         & 0.0\%        & 0.0\%         & 0.0\%        & 0.0\%         & 0.0\%          & 79.5\%        & 0.0\%          \\ \hline
\textbf{truck}      & 0.0\%                                                         & 20.1\%                                                          & 0.0\%         & 0.0\%        & 0.0\%         & 0.0\%        & 0.0\%         & 0.0\%          & 0.0\%         & 79.9\%         \\ \hline
\end{tabular}%
}
\end{table}

\section{Experimental Setup}

\begin{figure}[h]%
\centering
\includegraphics[width=0.95\textwidth]{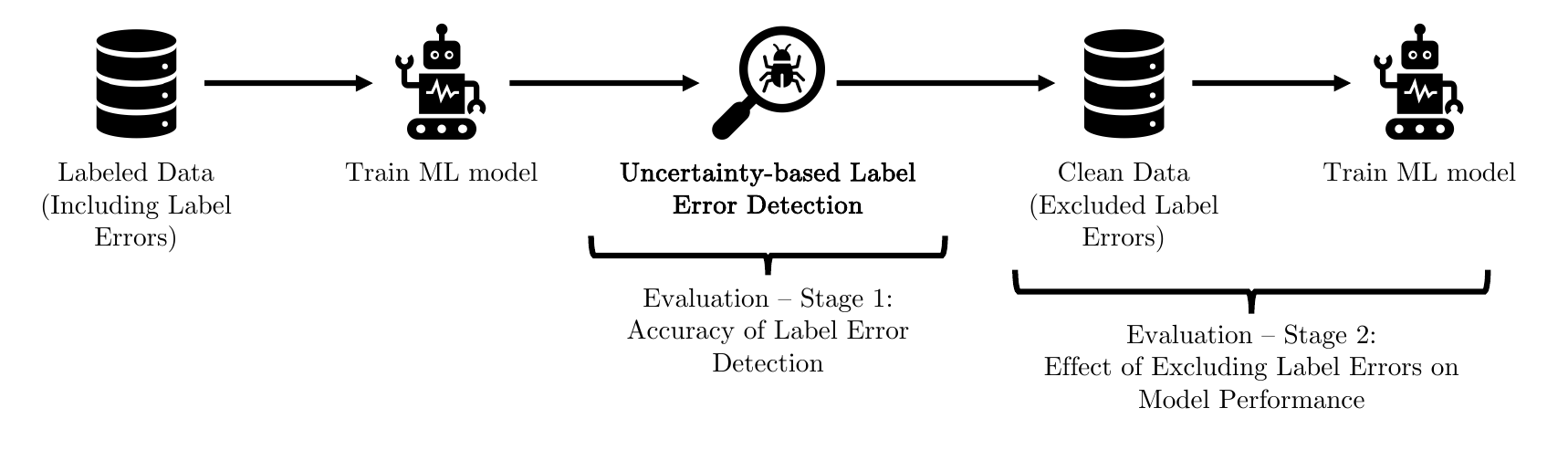}
\caption[Overview of the procedure of uncertainty quantification-based label error detection]{Simplified overview of the procedure of uncertainty quantification-based label error detection. }
\label{fig:approach}
\end{figure}

In our experiments, we focus on image classification due to its comparability with existing approaches, its practical relevance, and computational feasibility. Specifically we make use of the prominent datasets MNIST \cite{lecun_mnist_2010}, CIFAR-10 \cite{CIFAR-10}, CIFAR-100 \cite{CIFAR-10}, and tiny-ImageNet \cite{le_tiny_2015}. 
We evaluate each algorithm with regard to its label error detection performance as part of stage~1 of our evaluation. 
In stage~2 of the evaluation, we then measure the effect of excluding label errors from the training data on the final model accuracy based on different algorithms.
We present a simplified overview of this procedure in Figure~\ref{fig:approach}.
During both stages, the state-of-the-art CL-PBNR algorithm is employed as a baseline \cite{northcutt_confident_2019}.

In the following, we will examine our two-stage evaluation procedure in greater detail.
First, we explain how we evaluate our algorithms with regard to the two evaluation criteria, label error detection performance and the effect on model accuracy, in a controlled setting.
Subsequently, we will introduce the datasets for evaluation.

\subsection{Performance Metrics}
To evaluate our algorithms with regard to label error detection performance and the effect on model accuracy, synthetic label noise must be generated and added to the datasets before training.
This is required because we can only measure the label error detection performance of our algorithms and compare the novel algorithms to the state-of-the-art if we have ground truth data on whether a label is correct or wrong. 
Furthermore, the noise rate $\tau$ must be controllable to investigate the effects of different noise rates on the algorithms' performance and the achieved accuracy after cleaning. 

Contrary to other studies that deal with very high synthetic noise rates \cite{gare_exploiting_2021,northcutt_confident_2019,han_co-teaching_2018}, in this work, we want to use realistic noise rates as they occur in real-world datasets. \cite{northcutt_pervasive_2021} state that the test sets of common benchmark datasets, such as ImageNet, or Quickdraw, contain up to about 10\% of label errors. 
Therefore, we use noise rates $\tau_1=0.05$, $\tau_2=0.1$, and  $\tau_3=0.2$ for the evaluation of our algorithms. 

In stage~1, we measure the label error detection performance of the different algorithms with the F1 score. Note that we additionally evaluated the area under precision-recall curve (AUPR) and the F0.5 score and found consistent results. For reasons of brevity, we follow recent work and report F1 score, precision, and recall.
The F1 score equals the harmonic mean of recall and precision and is defined in the following equation.
\begin{equation}
\label{eq:f1}
    \mathrm{F1}=2 \cdot \frac{\mathrm{Precision} \cdot \mathrm{Recall}}{\mathrm{Precision} + \mathrm{Recall}}
\end{equation}
The precision reflects how many of the selected samples, i.e., true positives (TP) and false positives (FP), are actually relevant. 
It is defined as 
\begin{equation}
    \mathrm{Precision} = \frac{TP}{TP+FP} \; .
\end{equation}
The recall reflects how many relevant items, i.e., TP and FN, are selected. 
It is defined as
\begin{equation}
    \mathrm{Recall} = \frac{TP}{TP+FN} \; .
\end{equation} 

\begin{figure}[h]%
\centering
\includegraphics[width=0.95\textwidth]{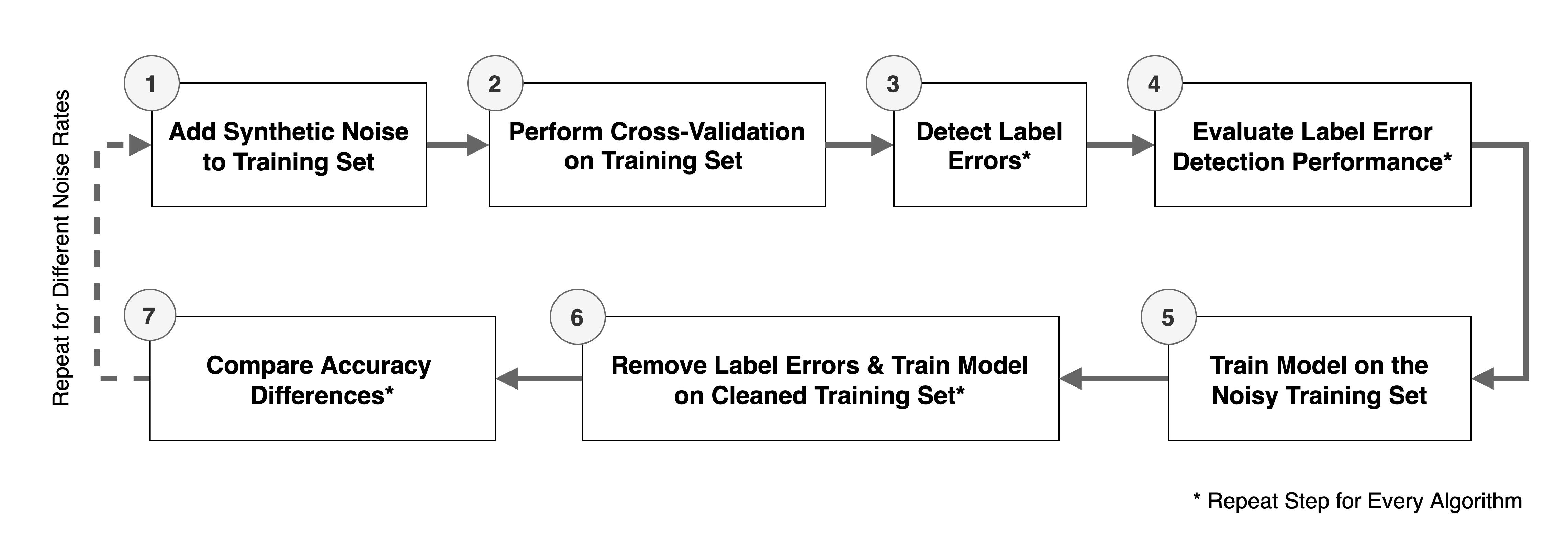}
\caption[In-depth evaluation procedure of the proposed algorithms for different noise rates]{In-depth evaluation procedure of the proposed algorithms for different noise rates.}
\label{fig:stage-2-flow}
\end{figure}

Apart from measuring the label error detection performance of our algorithms, we also want to assess their practical implications in terms of the effect on model accuracy after label error removal.
Consistent with the approach of \cite{northcutt_confident_2019}, we train a model on the noisy training set and another model on the cleaned training set before evaluating both on the unaltered test set. 
This allows us to determine the effect of cleaned data on model accuracy for each algorithm.
We will refer to the accuracy before cleaning as \textit{noisy accuracy}, and the accuracy after cleaning as \textit{clean accuracy}. 
We describe our evaluation approach based on seven subsequent phases, which are explained in Figure~\ref{fig:stage-2-flow}.

First, label errors are synthetically generated and added to the training set. 
Second, out-of-sample predictions are computed from the noisy training set via k-fold cross-validation. 
Following \cite{northcutt_confident_2019}, we conduct a stratified four-fold cross-validation. The predictions on each hold-out set are then combined to get the out-of-sample predictions for the entire noisy training set. 
Third, label errors are detected by the various algorithms. 
Fourth, each algorithm's label error detection performance is evaluated regarding recall, precision, and F1 score. 
Fifth, a model is trained on the noisy training set without cross-validation and evaluated on the unaltered test set to obtain the noisy accuracy. 
Sixth, for each algorithm, the label errors identified are automatically removed from the noisy training set, and a model is trained on the cleaned training set without cross-validation. The model is then evaluated on the unaltered test set to obtain the clean accuracy. 
Seventh, the clean accuracies achieved by our proposed algorithms are compared to the clean accuracy of the baseline (CL-PBNR) before then comparing the clean accuracies of all algorithms to the noisy accuracy.
All stages are repeated for the noise rates $\tau_1=0.05$, $\tau_2=0.1$, and $\tau_3=0.2$.

\textbf{Measuring Accuracy Differences:} 
In stage~2, we also measure the differences in test set accuracy before and after cleaning.
To calculate the noisy accuracy before label error removal, we train the dataset-specific model from scratch (without cross-validation) on the noisy training set and calculate the model's accuracy on the unaltered test set. 
For each algorithm, we then remove the identified label errors from the noisy training set and train the model again from scratch on the cleaned training set. 
Afterward, we evaluate the model's accuracy on the unaltered test set to obtain the clean accuracy for the respective algorithm. 
Lastly, we calculate the difference in accuracy between the model trained on the noisy and on the cleaned training set to measure the practical benefits of each algorithm. 
We use the same dataset-specific training settings as described above in the \textit{noise generation} paragraph to train the model from scratch on the noisy and cleaned training sets.
Note that we conduct a pilot study, where we evaluate all algorithms on the respective test sets of the datasets in order to determine which algorithms outperform CL-PBNR. These algorithms are then selected for an in-depth evaluation in order to reduce overall computation.

\subsection{Datasets}

\begin{figure*}[h]
    \centering
     \begin{subfigure}[t]{0.48\textwidth}
          \centering
\includegraphics[width=0.9\textwidth]{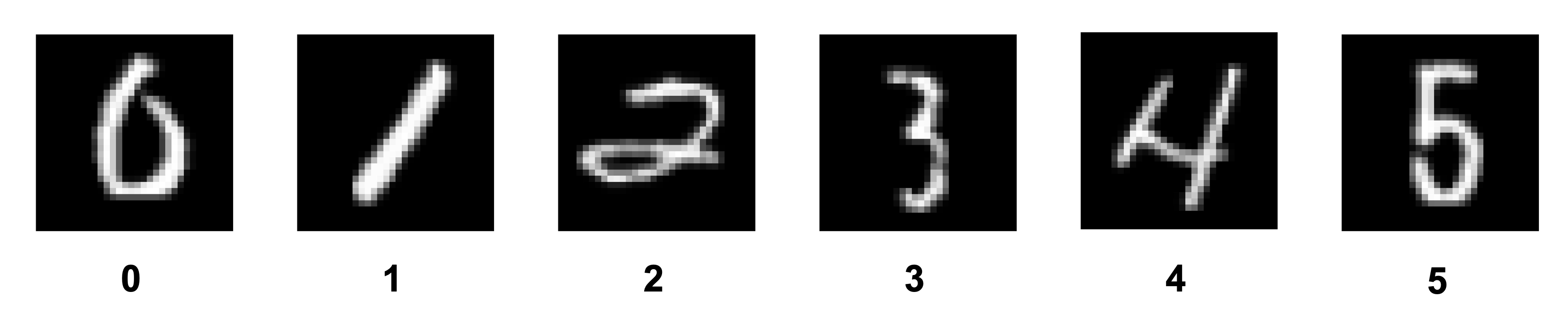}
    \caption{MNIST.}
    \label{fig:MNIST-example}
     \end{subfigure}
    \centering
        \begin{subfigure}[t]{0.48\textwidth}
         \centering
\includegraphics[width=0.9\textwidth]{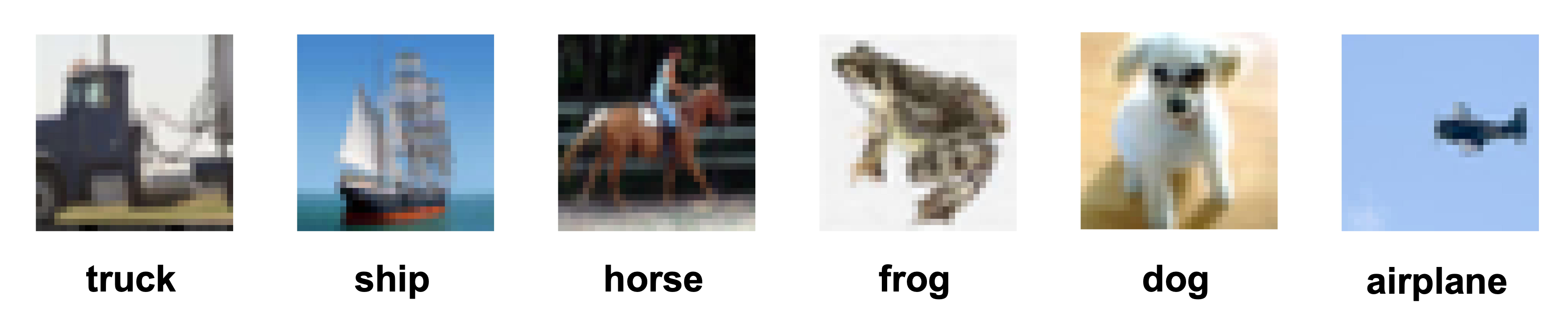}
    \caption{CIFAR-10.}
    \label{fig:CIFAR-10-example}
     \end{subfigure}     
     \begin{subfigure}[t]{0.48\textwidth}
         \centering
\includegraphics[width=0.9\textwidth]{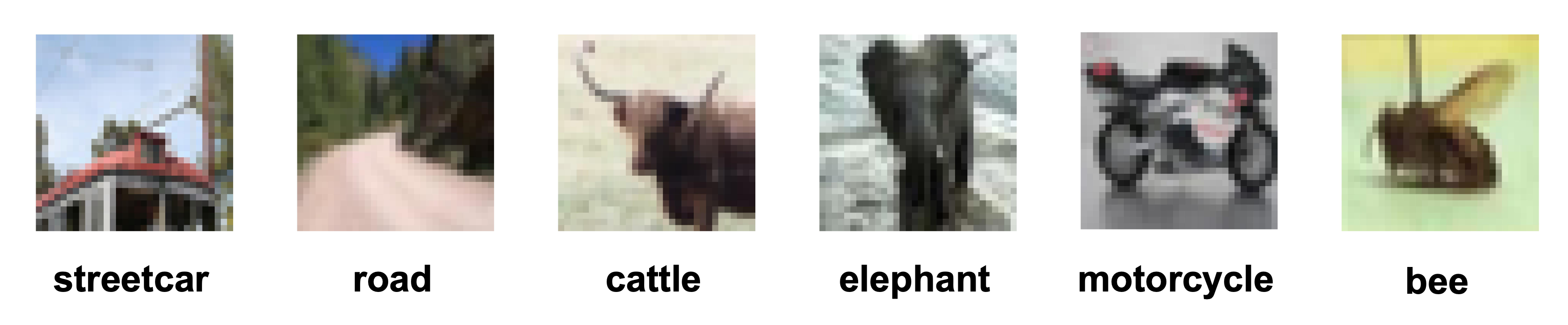}
    \caption{CIFAR-100.}
    \label{fig:CIFAR-100-example}
     \end{subfigure}     
\begin{subfigure}[t]{0.48\textwidth}
         \centering
\includegraphics[width=0.9\textwidth]{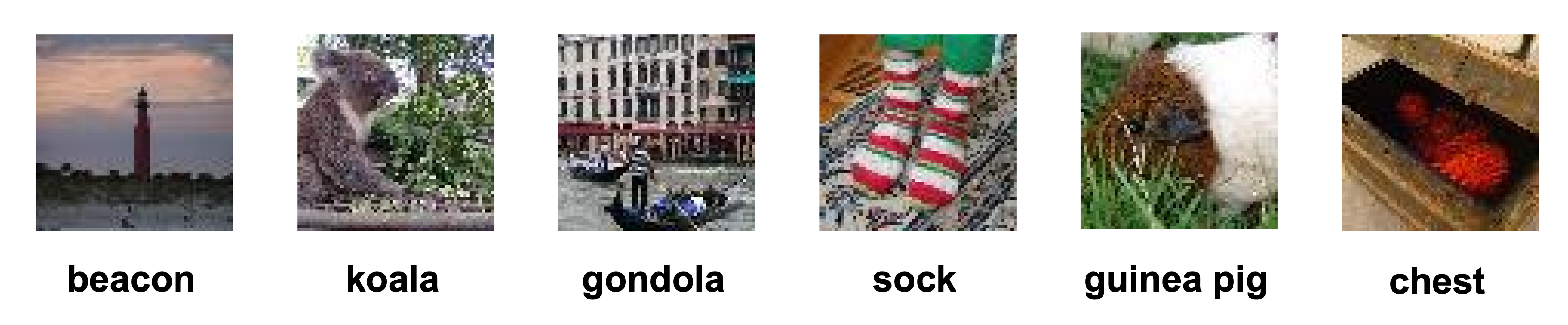}
    \caption{Tiny-Imagenet.}
    \label{fig:Tiny-Imagenet-example}
     \end{subfigure}     
     \caption{Example visualizations from each considered dataset.}
\end{figure*}

The MNIST dataset published by \cite{lecun_mnist_2010} is a prominent {ML} benchmark dataset of handwritten black and white digits ranging from zero to nine. It consists of 60,000 images for training and 10,000 images for testing. Examples of the MNIST dataset are depicted in Figure~\ref{fig:MNIST-example}.
The CIFAR-10 dataset published by \cite{CIFAR-10} is another commonly used {ML} benchmark dataset of colored images from ten different classes and consists of 50,000 training and 10,000 testing samples. Examples of the CIFAR-10 dataset can be seen in Figure~\ref{fig:CIFAR-10-example}.
The CIFAR-100 dataset, also published by \cite{CIFAR-10}, consists of colored images from 100 different classes and contains 50,000 training samples and 10,000 testing samples. Thus, for each class, there are only 500 training samples. Examples of the CIFAR-100 dataset can be seen in Figure~\ref{fig:CIFAR-100-example}.
The Tiny-ImageNet dataset published by \cite{le_tiny_2015} is a reduced version of the popular ImageNet classification challenge. 
The dataset consists of colored images from 200 different classes and has 100,000 training and 10,000 testing samples. 
With just 500 training images per class, this dataset has, together with the CIFAR-100 dataset, the lowest number of training images per class and the highest number of classes. In Figure~\ref{fig:Tiny-Imagenet-example}, we depict examples of the Tiny-ImageNet dataset.

\subsection{Implementation Details}

In the following, we describe the pre-training, the noise generation, cross-validation, as well as hardware and frameworks. 

\textbf{Pre-Training} For pre-training on the original training sets, we use a modified version of the VGG \cite{simonyan_very_2014} network as well as ResNet \cite{he2016deep} architecture. Specifically, we use a VGG-11 model for the MNIST dataset and ResNet-50 models for CIFAR-10, CIFAR-100, and Tiny-ImageNet. 
We adapt both model architectures towards uncertainty quantification by adding five dropout layers after the five most inner encoder layers, following \cite{kendall_bayesian_2015}.
The dropout probability of those layers is set to the standard value of 0.5 \cite{srivastava_dropout_2014}. 
For the pre-training on the MNIST training set, we train the modified VGG-11 model over 15 epochs with a batch size of 256 and a learning rate of 0.01.
For the pre-training on the CIFAR-10, CIFAR-100, and Tiny-ImageNet datasets, we run three modified ResNet-50 models over 350 epochs with a batch size of 256. The learning rate scheduler reduces the initial learning rate of 0.1 after 150 epochs to 0.01 and, finally, after 250 epochs to 0.001.
For all datasets, we split the non-test data into 80\% training and 20\% validation data.
During the training of all models, we drop the last partial batch of each epoch. 
According to \cite{northcutt_confident_2019}, this improves stability by avoiding weight updates from just a few noisy samples. 

\textbf{Noise Generation} Before obtaining out-of-sample predicted probabilities from the MNIST, CIFAR-10, CIFAR-100, and Tiny-ImageNet datasets, we add synthetic asymmetric label noise to each dataset based on the method proposed above. 
For the noise generation, we leverage the previously pretrained models. After pre-training, we let each model predict on its test set and use the softmax probabilities to calculate the class-specific similarity scores and flipping probabilities per dataset. 
Finally, using the calculated similarity scores, we generate the noise transition matrices for the three different noise rates $\tau_1=0.05$, $\tau_2=0.1$, and $\tau_3=0.2$ for each dataset. 

\textbf{Cross-Validation} After the label noise has been added to each training set, we conduct a four-fold cross-validation to obtain out-of-sample probabilities.  
For the cross-validation, we use the same model and training settings per dataset as described above. 
To obtain out-of-sample predicted {MCD} probabilities during the cross-validation, we conduct $F=5$ forward passes. 
After the out-of-sample predicted softmax and {MCD} probabilities are obtained via the cross-fold validation, we identify label errors in the training set using the different algorithms and measure their label error detection performance. 

\textbf{Hardware and Frameworks} 
Every model is trained on an NVIDIA Tesla V100. All code is developed in \textit{PyTorch} \cite{Pytorch2019}, and \textit{PyTorch Lightning} \cite{falcon_pytorch_2019}, which is a high-level interface for \textit{PyTorch} simplifying many repetitive tasks, like monitoring.

\section{Experimental Results}
In this section, we report the results and the key findings from our experiments. As introduced in Figure~\ref{fig:approach}, we holistically evaluate the proposed algorithms in two stages. In the first stage, we assess the accuracy of detecting label errors within popular image classification datasets. In the second stage, we then examine the effect of excluding these potential label errors from the training datasets on the final model performance. Across both stages, we compare all proposed algorithms with the state-of-the-art label error detection represented by the CL-PBNR algorithm. We report our results across different levels of label errors in the datasets to gain additional insights into the effects of different amounts of label errors on performance.

\subsection{Evaluation Stage 1: Accuracy of Label Error Detection}
We report the accuracy in detecting label errors in terms of F1 scores, precision, and recall of all proposed algorithms alongside the state-of-the-art algorithm of CL-PBNR in Table~\ref{tab:stage-1-absolute}. The experiments consistently demonstrate that all proposed uncertainty-based algorithms outperform the current state-of-the-art in label error detection for all considered datasets and across noise rates. 
The mean performance over CL-PBNR increases by 2.1pp for MNIST, 14.3pp for CIFAR-10, 8.2pp for CIFAR-100, and 6.6pp for Tiny-ImageNet. 
Additionally, we make several important observations. 

\begin{table}[ht]
\caption[Accuracy in identifying label errors for uncertainty quantification-based label error detection compared to the baseline]{Evaluation stage~1: Accuracy in identifying label errors for uncertainty quantification-based label error detection compared to the baseline, \textit{prune by noise rate} algorithm CL-PBNR \cite{northcutt_confident_2019}.}
\centering
 \resizebox{\textwidth}{!}{%
\begin{tabular}{l|cccc|ccc|ccc}
\toprule
Measure                     & \multicolumn{4}{c|}{F1}                                                                    & \multicolumn{3}{c|}{Precision}                      & \multicolumn{3}{c}{Recall}                          \\
Noise Rate $\tau$           & 0.05            & 0.1             & \multicolumn{1}{c|}{0.2}             & Mean       & 0.05            & 0.1             & 0.2             & 0.05            & 0.1             & 0.2             \\ 
\midrule
\multicolumn{11}{c}{\textbf{MNIST}} \\
\midrule
Baseline (CL-PBNR)                     & 93.9\%          & 94.2\%          & \multicolumn{1}{c|}{94.4\%}          & 94.2\%          & 92.6\%          & 94.4\%          & 94.9\%          & 95.3\%          & 93.9\%          & 93.9\%          \\
CL-MCD                      & 95.4\%          & 95.2\%          & \multicolumn{1}{c|}{95.0\%}          & 95.2\%          & 96.2\%          & 97.4\%          & 97.5\%          & 94.7\%          & 93.2\%          & 92.7\%          \\
CL-MCD-E                  & 95.2\%          & 94.6\%          & \multicolumn{1}{c|}{92.3\%}          & 94.0\%          & 94.9\%          & 97.1\%          & 97.4\%          & 95.5\%          & 92.2\%          & 87.7\%          \\
CL-MCD-Ens             & 95.4\%          & 95.5\%          & \multicolumn{1}{c|}{95.0\%}          & 95.3\%          & 95.3\%          & 96.7\%          & 97.2\%          & 95.5\%          & 94.2\%          & 93.0\%          \\
Alg. Ens (Agree.=2) & \textbf{96.1\%} & \textbf{96.2\%} & \multicolumn{1}{c|}{\textbf{95.9\%}} & \textbf{96.1\%} & 95.8\%          & 97.4\%          & 97.1\%          & \textbf{96.4\%} & \textbf{95.0\%} & \textbf{94.8\%} \\
Alg. Ens (Agree.=3) & 95.8\%          & 94.9\%          & \multicolumn{1}{c|}{93.4\%}          & 94.7\%          & \textbf{97.2\%} & \textbf{98.4\%} & \textbf{98.7\%} & 94.4\%          & 91.7\%          & 88.7\%          \\ 
\midrule
\multicolumn{11}{c}{\textbf{CIFAR-10}} \\
\midrule
Baseline (CL-PBNR)                     & 49.3\%          & 56.1\%          & \multicolumn{1}{c|}{60.3\%}          & 55.2\%          & 33.7\%          & 41.1\%          & 47.7\%          & 92.0\%          & 88.4\%          & 82.0\%          \\
CL-MCD                      & 61.2\%          & 63.9\%          & \multicolumn{1}{c|}{63.4\%}          & 62.8\%          & 46.7\%          & 51.0\%          & 52.6\%          & 88.8\%          & 85.7\%          & 80.0\%          \\
CL-MCD-E                  & \textbf{67.6\%} & \textbf{71.5\%} & \multicolumn{1}{c|}{\textbf{69.4\%}} & \textbf{69.5\%} & \textbf{55.7\%} & \textbf{62.5\%} & \textbf{62.0\%} & 86.0\%          & 83.5\%          & 78.7\%          \\
CL-MCD-Ens             & 53.0\%          & 59.2\%          & \multicolumn{1}{c|}{62.3\%}          & 58.2\%          & 37.1\%          & 44.1\%          & 49.7\%          & \textbf{92.8\%} & \textbf{89.7\%} & \textbf{83.3\%} \\
Alg. Ens (Agree.=2) & 61.2\%          & 64.4\%          & \multicolumn{1}{c|}{64.4\%}          & 63.3\%          & 46.5\%          & 51.3\%          & 53.1\%          & 89.5\%          & 86.7\%          & 81.8\%          \\
Alg. Ens (Agree.=3) & 64.0\%          & 67.1\%          & \multicolumn{1}{c|}{66.1\%}          & 65.7\%          & 50.8\%          & 56.2\%          & 57.4\%          & 86.5\%          & 83.4\%          & 77.9\%          \\ 
\midrule
\multicolumn{11}{c}{\textbf{CIFAR-100}} \\
\midrule
Baseline (CL-PBNR)                     & 24.4\%          & 38.6\%          & \multicolumn{1}{c|}{53.4\%}          & 38.8\%          & 14.1\%          & 24.5\%          & 37.9\%          & 89.6\%          & 90.8\%          & 90.6\%          \\
CL-MCD                      & 28.5\%          & 42.7\%          & \multicolumn{1}{c|}{56.7\%}          & 42.6\%          & 17.1\%          & 28.3\%          & 41.9\%          & 85.8\%          & 87.1\%          & 87.3\%          \\
CL-MCD-E                  & 32.5\%          & 46.4\%          & \multicolumn{1}{c|}{59.0\%}          & 46.0\%          & 20.5\%          & 32.7\%          & 46.7\%          & 77.4\%          & 79.9\%          & 80.3\%          \\
CL-MCD-Ens             & 25.1\%          & 39.4\%          & \multicolumn{1}{c|}{54.2\%}          & 39.6\%          & 14.6\%          & 25.0\%          & 38.2\%          & \textbf{92.2\%} & \textbf{93.6\%} & \textbf{93.7\%} \\
Alg. Ens (Agree.=2) & 29.3\%          & 43.5\%          & \multicolumn{1}{c|}{57.4\%}          & 43.4\%          & 17.6\%          & 29.0\%          & 42.5\%          & 85.7\%          & 87.2\%          & 88.2\%          \\
Alg. Ens (Agree.=3) & \textbf{33.7\%} & \textbf{47.8\%} & \multicolumn{1}{c|}{\textbf{59.6\%}} & \textbf{47.0\%} & \textbf{21.9\%} & \textbf{34.8\%} & \textbf{48.9\%} & 73.4\%          & 76.2\%          & 76.2\%          \\ 
\midrule
\multicolumn{11}{c}{\textbf{Tiny-ImageNet}} \\
\midrule
Baseline (CL-PBNR)                     & 22.5\%          & 37.4\%          & \multicolumn{1}{c|}{55.1\%}          & 38.3\%          & 13.0\%          & 24.0\%          & 40.7\%          & 82.5\%          & 84.2\%          & 85.2\%          \\
CL-MCD                      & 24.6\%          & 39.5\%          & \multicolumn{1}{c|}{57.5\%}          & 40.5\%          & 14.5\%          & 26.0\%          & 43.7\%          & 81.2\%          & 81.9\%          & 84.2\%          \\
CL-MCD-E                  & 27.2\%          & 43.0\%          & \multicolumn{1}{c|}{\textbf{61.3\%}} & 43.8\%          & 16.5\%          & 29.6\%          & 49.1\%          & 76.6\%          & 78.7\%          & 81.7\%          \\
CL-MCD-Ens             & 23.8\%          & 38.8\%          & \multicolumn{1}{c|}{57.1\%}          & 39.9\%          & 13.8\%          & 25.1\%          & 42.3\%          & \textbf{85.3\%} & \textbf{85.8\%} & \textbf{87.8\%} \\
Alg. Ens (Agree.=2) & 25.6\%          & 40.9\%          & \multicolumn{1}{c|}{59.0\%}          & 41.8\%          & 15.2\%          & 27.2\%          & 45.2\%          & 81.1\%          & 82.5\%          & 85.0\%          \\
Alg. Ens (Agree.=3) & \textbf{29.0\%} & \textbf{44.5\%} & \multicolumn{1}{c|}{61.1\%}          & \textbf{44.9\%} & \textbf{18.4\%} & \textbf{32.5\%} & \textbf{52.6\%} & 67.9\%          & 70.4\%          & 72.9\%          \\ 
\bottomrule
\end{tabular}%
}
\label{tab:stage-1-absolute}
\end{table}

\begin{figure*}[h]
    \centering
     \begin{subfigure}[t]{0.48\textwidth}
          \centering
\includegraphics[width=\textwidth]{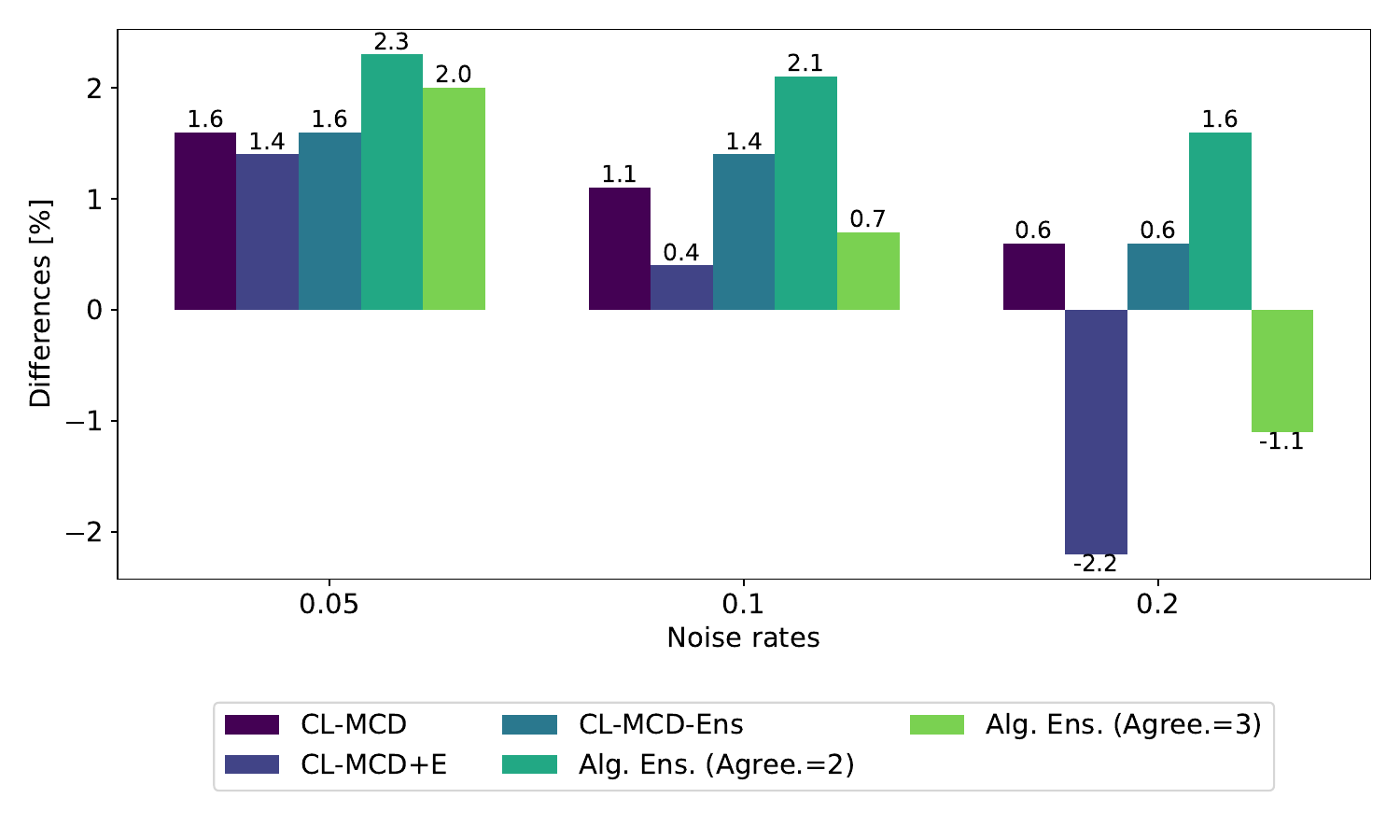}
    \caption{MNIST}
     \end{subfigure}
    \centering
        \begin{subfigure}[t]{0.48\textwidth}
         \centering
\includegraphics[width=\textwidth]{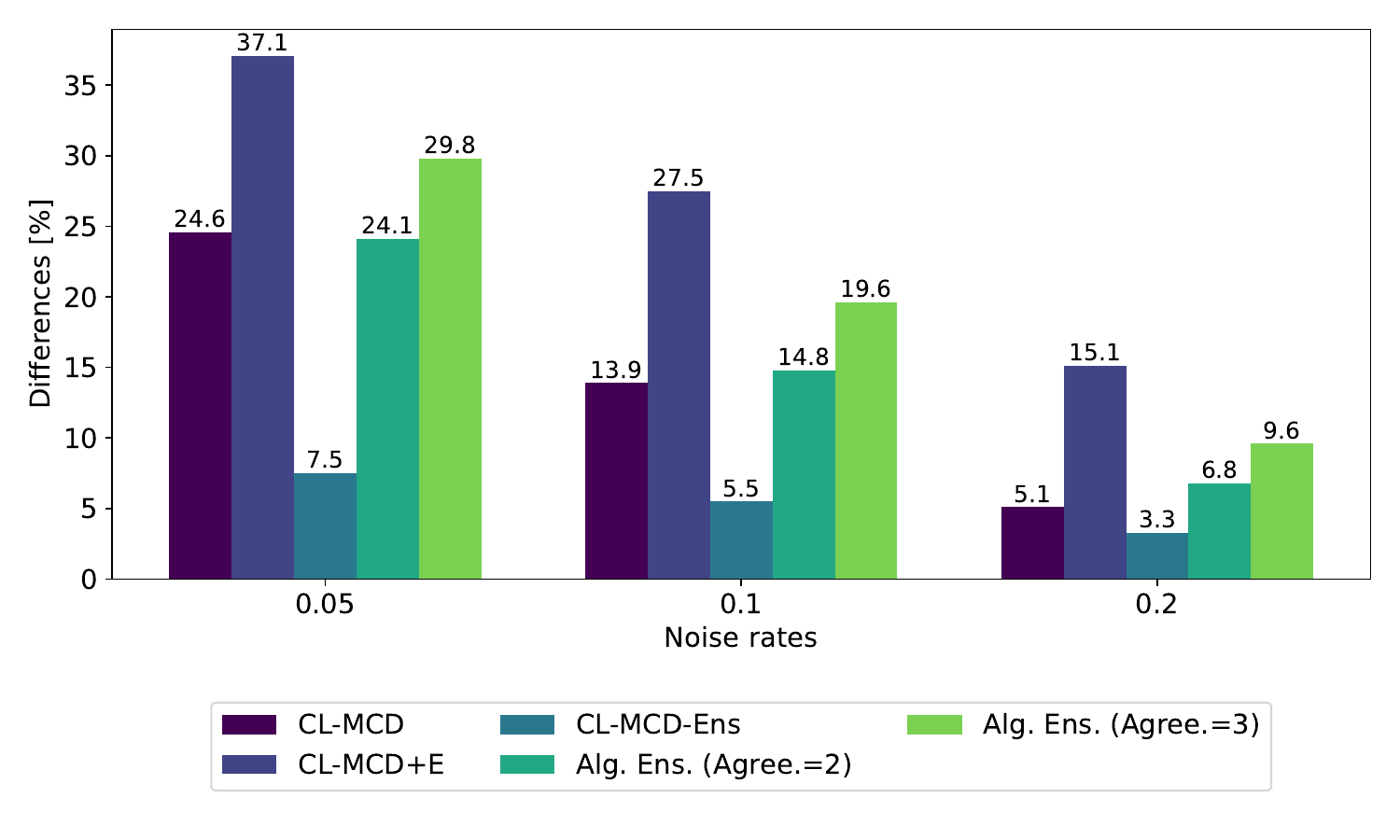}
    \caption{CIFAR-10}
     \end{subfigure}     
     \begin{subfigure}[t]{0.48\textwidth}
         \centering
\includegraphics[width=\textwidth]{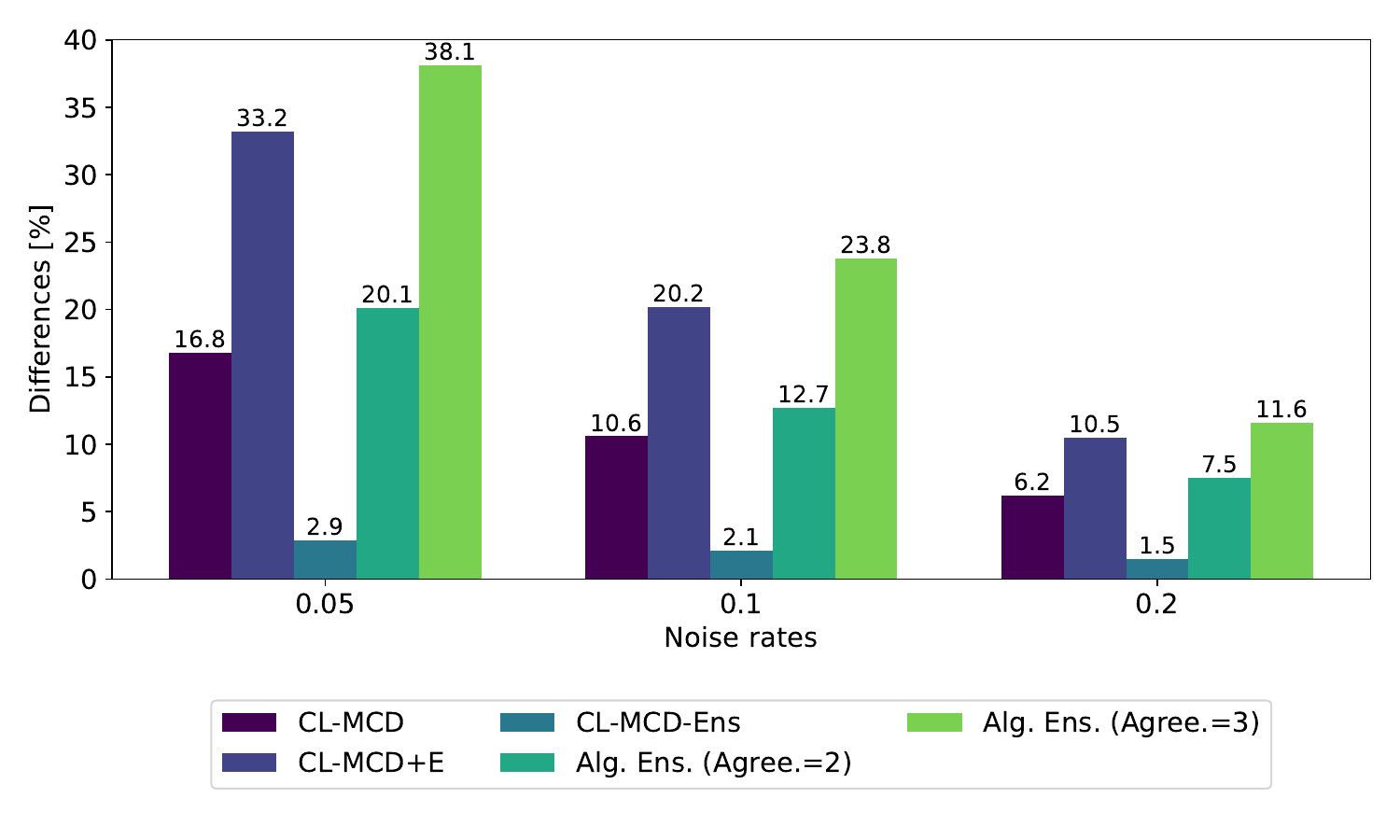}
    \caption{CIFAR-100}
     \end{subfigure}     
\begin{subfigure}[t]{0.48\textwidth}
         \centering
\includegraphics[width=\textwidth]{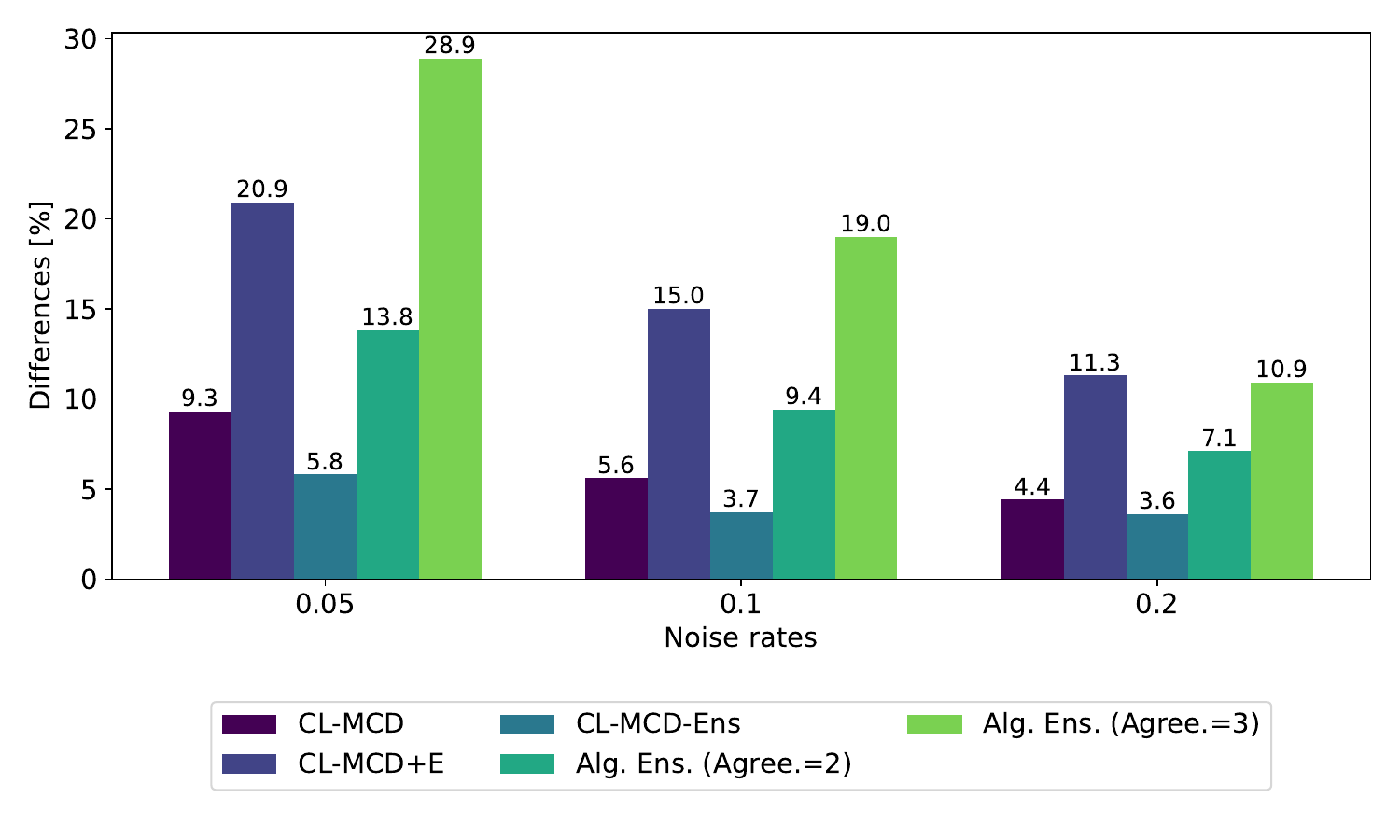}
    \caption{Tiny-Imagenet}
     \end{subfigure}     
     \caption{Relative comparison between the proposed algorithms and the CL-PBNR algorithm in terms of increases or decreases in the F1 score.}
     \label{fig:stage-1-relative}
\end{figure*}

First, our experiments suggest that it is beneficial to leverage MCD probabilities instead of softmax probabilities in general. Utilizing MCD probabilities improves the performances even when we do not calculate additional uncertainty measures, like entropy, on top of the probabilities (see the performance of CL-MCD). For example, the mean performance over CL-PBNR increases by 1.0pp for MNIST, 7.6pp for CIFAR-10, 3.8pp for CIFAR-100, and 2.2pp for Tiny-ImageNet. 
Second, the performance consistently increases when we leverage entropy as a measure of uncertainty on top of MCD probabilities. This underlines the importance of quantifying model uncertainty, especially for lower noise rates where the relative performance improvement over the state-of-the-art is particularly pronounced. We present these relative improvements in Figure~\ref{fig:stage-1-relative}.
Third, combining different algorithms in an ensemble can additionally improve the performances by leveraging different capabilities. This behavior is particularly pronounced for more complex datasets like CIFAR-100 and Tiny-ImageNet.

The experiments reveal that the performance improvements over the state-of-the-art are especially large for more complex datasets. For CIFAR-10, CIFAR-100, and Tiny-ImageNet, our best-performing algorithms CL-MCD-E and the Algorithm Ensemble (Agreement = 3) improve over the state-of-the-art by at least 10\% across all considered noise rates (see Figure~\ref{fig:stage-1-relative}). The best-performing algorithms outperform the state-of-the-art by up to 9.3pp in Table~\ref{tab:stage-1-absolute} (or 38.1\%, see CIFAR-100, $\tau=0.05$ in Figure~\ref{fig:stage-1-relative}). 
We further note that our uncertainty-based algorithms perform particularly strong compared to the state-of-the-art when the relative share of label errors on the full dataset is small (e.g., $\tau=0.05$). This observation is especially relevant since recent work has found that the share of label errors on real-world datasets ranges between 1\% and 10\% of the dataset size \cite{northcutt_pervasive_2021}. Thus, the advantages of our algorithms enlarge when noise rates become more realistic. 

The improvement in the detection of label errors (i.e., increasing F1 scores) of the proposed algorithms can be attributed to significant increases in the precision scores accompanied by smaller improvements in the recall. In Table~\ref{tab:stage-1-absolute}, we observe that the shortcomings of CL-PBNR lie in the precision with scores below 50\% for CIFAR-10, CIFAR-100, and Tiny-ImageNet, while the recall of CL-PBNR is already consistently over 80\%. All our algorithms increase both precision and recall. The precision improves by up to 22pp (or 65.28\%, see CIFAR-10 results at $\tau=0.05$) while the recall increases slightly by up to 2.8 (or 3.39\%, see Tiny-ImageNet results at $\tau=0.05$).

\subsection{Evaluation Stage 2: Effect of Excluding Label Errors on Model Performance}
In the second stage of our experiments, we assess the effect of excluding the previously identified label errors from the training datasets on the performance of the ML models. For these experiments, potential label errors are only removed from the training data while the test datasets remain unchanged. 
Overall, for all datasets besides Tiny-Imagenet, excluding label errors from the training dataset improves the model performance with all proposed algorithms and CL-PBNR baseline. 
The mean performance increases based on excluding label errors with the proposed algorithms compared to training on the original training datasets, amounting to up to 0.2pp for MNIST, 7.1pp for CIFAR-10, and 3.5pp for CIFAR-100 as reported in Table~\ref{tab:stage-2-absolute}. 

\begin{table}[ht]
\caption[Final model accuracy after removing label errors from the training datasets compared to the baseline]{Evaluation stage~2: Final model accuracy after removing label errors from the training datasets. Test sets remain unchanged.\footnote{As a baseline, the rows named \textit{original dataset} refer to the performance of models that have been trained on the datasets that still include label errors.}}
\centering
 \resizebox{\textwidth}{!}{%
\begin{tabular}{l|cccc|ccc}
\toprule
Metric                     & \multicolumn{4}{c|}{Accuracy}                                                              & \multicolumn{3}{c}{\# of potential label errors} \\
Noise Rate $\tau$           & 0.05            & 0.1             & \multicolumn{1}{c|}{0.2}             & Mean       & 0.05           & 0.1            & 0.2            \\ 
\midrule
\multicolumn{8}{c}{\textbf{MNIST}}  \\                          
\midrule
Original dataset                       & 99.0\%          & 98.9\%          & \multicolumn{1}{c|}{98.5\%}          & 98.8\%          & 0              & 0              & 0              \\
CL-PBNR                     & 98.9\%          & 98.8\%          & \multicolumn{1}{c|}{98.5\%}          & 98.7\%          & 3,089          & 5,970          & 11,879         \\
CL-MCD                      & \textbf{99.3\%} & 98.9\%          & \multicolumn{1}{c|}{98.7\%}          & \textbf{99.0\%} & 2,954          & 5,744          & 11,405         \\
CL-MCD-E                  & 99.2\%          & 98.8\%          & \multicolumn{1}{c|}{\textbf{99.0\%}} & \textbf{99.0\%} & 3,019          & 5,700          & 10,808         \\
CL-MCD-Ensemble             & 99.0\%          & \textbf{99.0\%} & \multicolumn{1}{c|}{98.1\%}          & 98.7\%          & 3,004          & 5,847          & 11,475         \\
Alg. Ens. (Agreement=2) & 99.2\%          & \textbf{99.0\%} & \multicolumn{1}{c|}{98.6\%}          & 98.9\%          & 3,020          & 5,850          & 11,718         \\
Alg. Ens. (Agreement=3) & 99.1\%          & 98.9\%          & \multicolumn{1}{c|}{98.7\%}          & 98.9\%          & 2,916          & 5,592          & 10,785         \\ 
\midrule
\multicolumn{8}{c}{\textbf{CIFAR-10}}  \\                          
\midrule
Original dataset                       & 89.2\%          & 84.3\%          & \multicolumn{1}{c|}{76.6\%}          & 83.4\%          & 0              & 0              & 0              \\
CL-PBNR                     & 91.9\%          & 90.2\%          & \multicolumn{1}{c|}{86.0\%}          & 89.4\%          & 6,820          & 10,754         & 17,182         \\
CL-MCD                      & 91.9\%          & 90.7\%          & \multicolumn{1}{c|}{86.2\%}          & 89.6\%          & 4,755          & 8,399          & 15,221         \\
CL-MCD-E                  & \textbf{92.6\%} & \textbf{91.1\%} & \multicolumn{1}{c|}{\textbf{87.8\%}} & \textbf{90.5\%} & 3,862          & 6,680          & 12,707         \\
CL-MCD-Ensemble             & 92.1\%          & 90.8\%          & \multicolumn{1}{c|}{86.4\%}          & 89.8\%          & 6,253          & 10,167         & 16,758         \\
Alg. Ens. (Agreement=2) & 92.4\%          & 90.8\%          & \multicolumn{1}{c|}{86.6\%}          & 89.9\%          & 4,806          & 8,451          & 15,412         \\
Alg. Ens. (Agreement=3) & 91.9\%          & 90.8\%          & \multicolumn{1}{c|}{86.4\%}          & 89.7\%          & 4,259          & 7,426          & 13,579         \\ 
\midrule
\multicolumn{8}{c}{\textbf{CIFAR-100}}  \\                          
\midrule
Original dataset                       & 69.6\%          & 66.3\%          & \multicolumn{1}{c|}{58.6\%}          & 64.8\%          & 0              & 0              & 0              \\
CL-PBNR                     & 68.0\%          & 66.1\%          & \multicolumn{1}{c|}{63.0\%}          & 65.7\%          & 15,860         & 18,544         & 23,942         \\
CL-MCD                      & 69.1\%          & 67.1\%          & \multicolumn{1}{c|}{63.8\%}          & 66.7\%          & 12,549         & 15,389         & 20,830         \\
CL-MCD-E                  & 70.3\%          & \textbf{69.3\%} & \multicolumn{1}{c|}{\textbf{65.2\%}} & \textbf{68.3\%} & 9,425          & 12,218         & 17,204         \\
CL-MCD-Ensemble             & 69.1\%          & 67.2\%          & \multicolumn{1}{c|}{62.8\%}          & 66.4\%          & 15,833         & 18,757         & 24,560         \\
Alg. Ens. (Agreement=2) & 68.0\%          & 68.0\%          & \multicolumn{1}{c|}{64.4\%}          & 66.8\%          & 12,148         & 15,049         & 20,737         \\
Alg. Ens. (Agreement=3) & \textbf{70.4\%} & \textbf{69.3\%} & \multicolumn{1}{c|}{65.0\%}          & 68.2\%          & 8,402          & 10,936         & 15,562         \\ 
\midrule
\multicolumn{8}{c}{\textbf{Tiny-Imagenet}}  \\                          
\midrule
Original dataset                       & \textbf{59.8\%} & \textbf{59.8\%} & \multicolumn{1}{c|}{\textbf{60.2\%}} & \textbf{59.9\%} & 0              & 0              & 0              \\
CL-PBNR                     & 56.4\%          & 56.0\%          & \multicolumn{1}{c|}{54.0\%}          & 55.5\%          & 31,669         & 35,040         & 41,883         \\
CL-MCD                      & 57.5\%          & 56.0\%          & \multicolumn{1}{c|}{54.5\%}          & 56.0\%          & 27,985         & 31,452         & 38,545         \\
CL-MCD-E                  & 57.7\%          & 57.2\%          & \multicolumn{1}{c|}{55.9\%}          & 56.9\%          & 23,174         & 26,587         & 33,257         \\
CL-MCD-Ensemble             & 57.4\%          & 56.4\%          & \multicolumn{1}{c|}{53.7\%}          & 55.8\%          & 30,818         & 34,252         & 41,528         \\
Alg. Ens. (Agreement=2) & 57.4\%          & 57.1\%          & \multicolumn{1}{c|}{55.6\%}          & 56.7\%          & 26,692         & 30,310         & 37,598         \\
Alg. Ens. (Agreement=3) & 58.8\%          & 57.2\%          & \multicolumn{1}{c|}{56.6\%}          & 57.5\%          & 18,422         & 21,647         & 27,705         \\ 
\bottomrule
\end{tabular}%
}
\label{tab:stage-2-absolute}
\end{table}

\begin{figure*}[h]
    \centering
     \begin{subfigure}[t]{0.48\textwidth}
          \centering
\includegraphics[width=\textwidth]{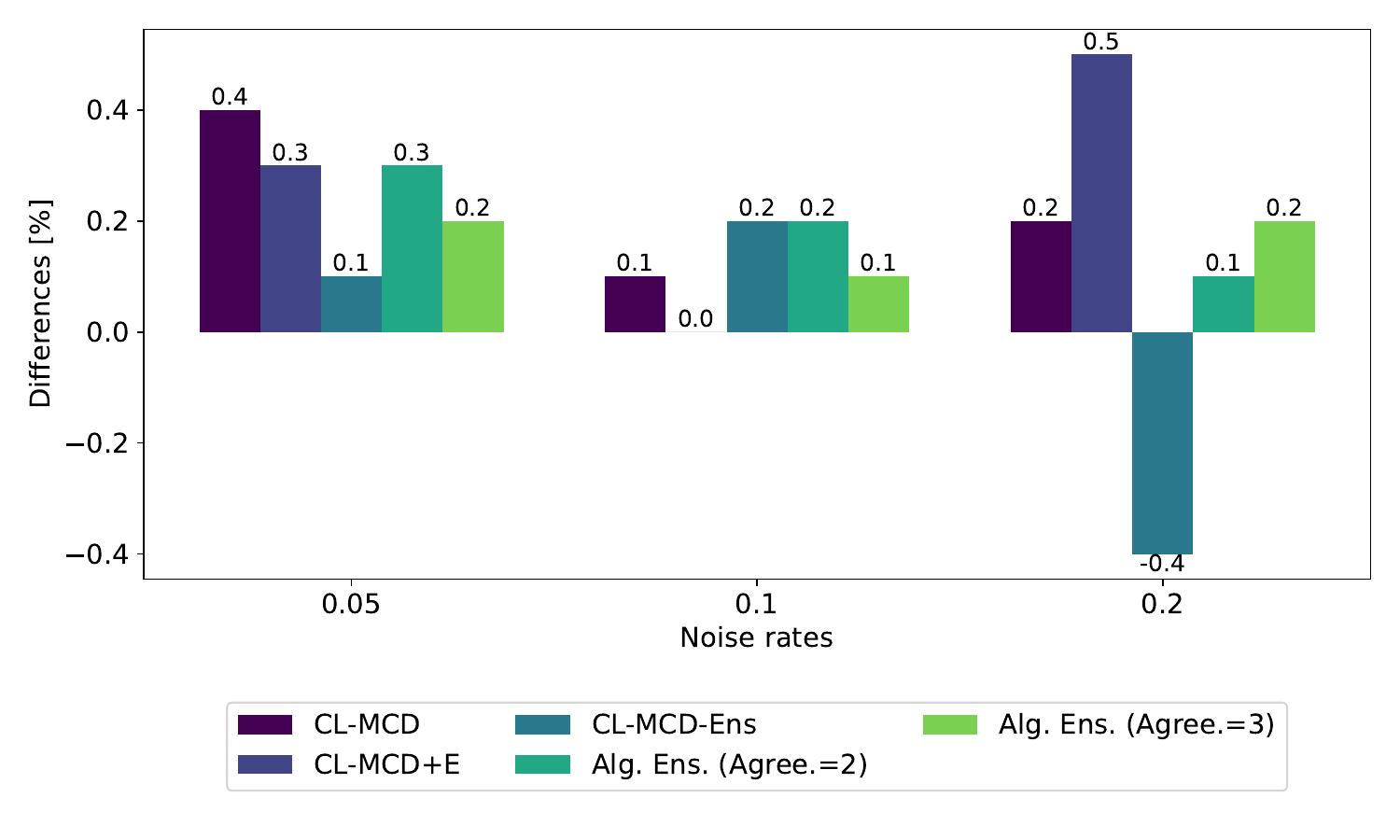}
    \caption{MNIST}
     \end{subfigure}
    \centering
        \begin{subfigure}[t]{0.48\textwidth}
         \centering
\includegraphics[width=\textwidth]{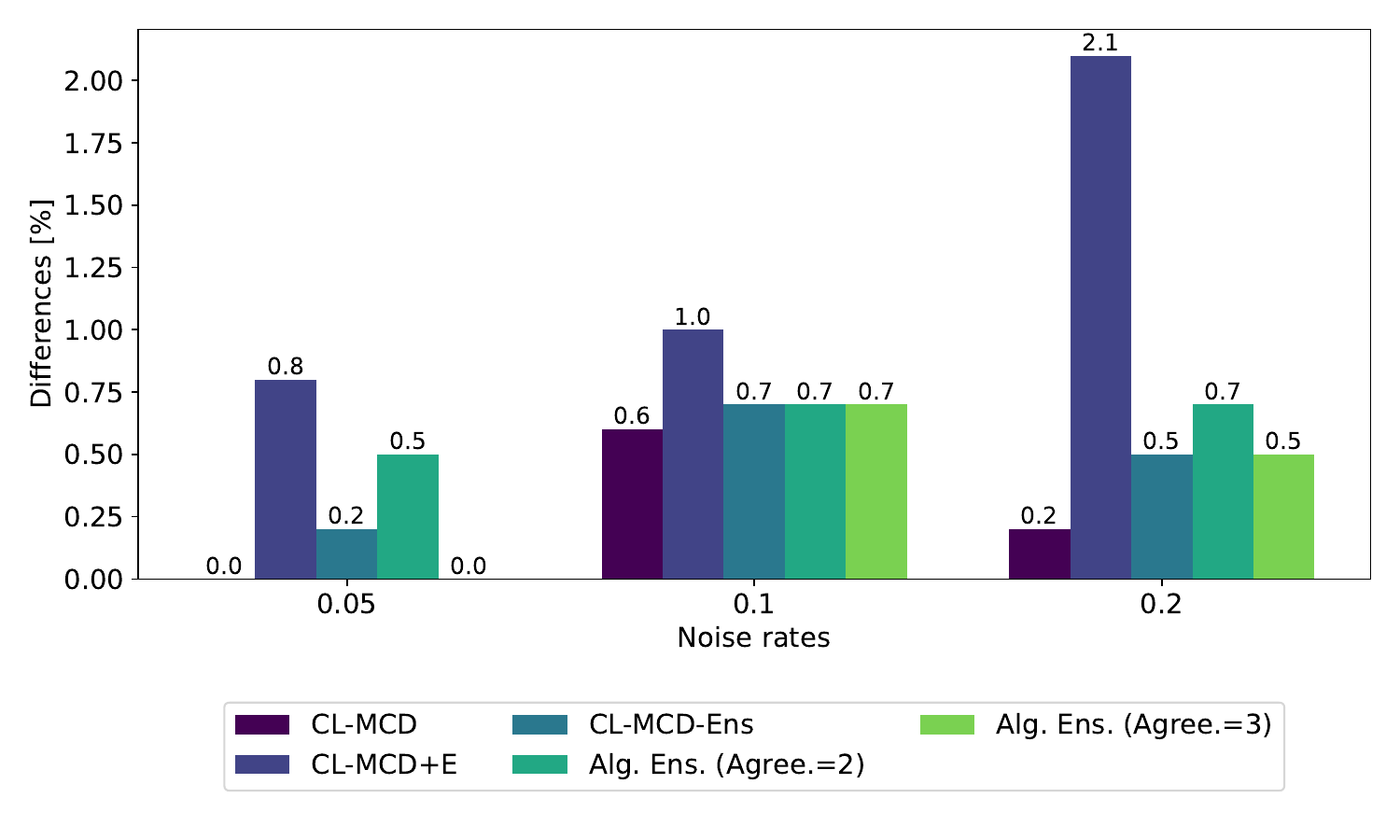}
    \caption{CIFAR-10}
     \end{subfigure}     
     \begin{subfigure}[t]{0.48\textwidth}
         \centering
\includegraphics[width=\textwidth]{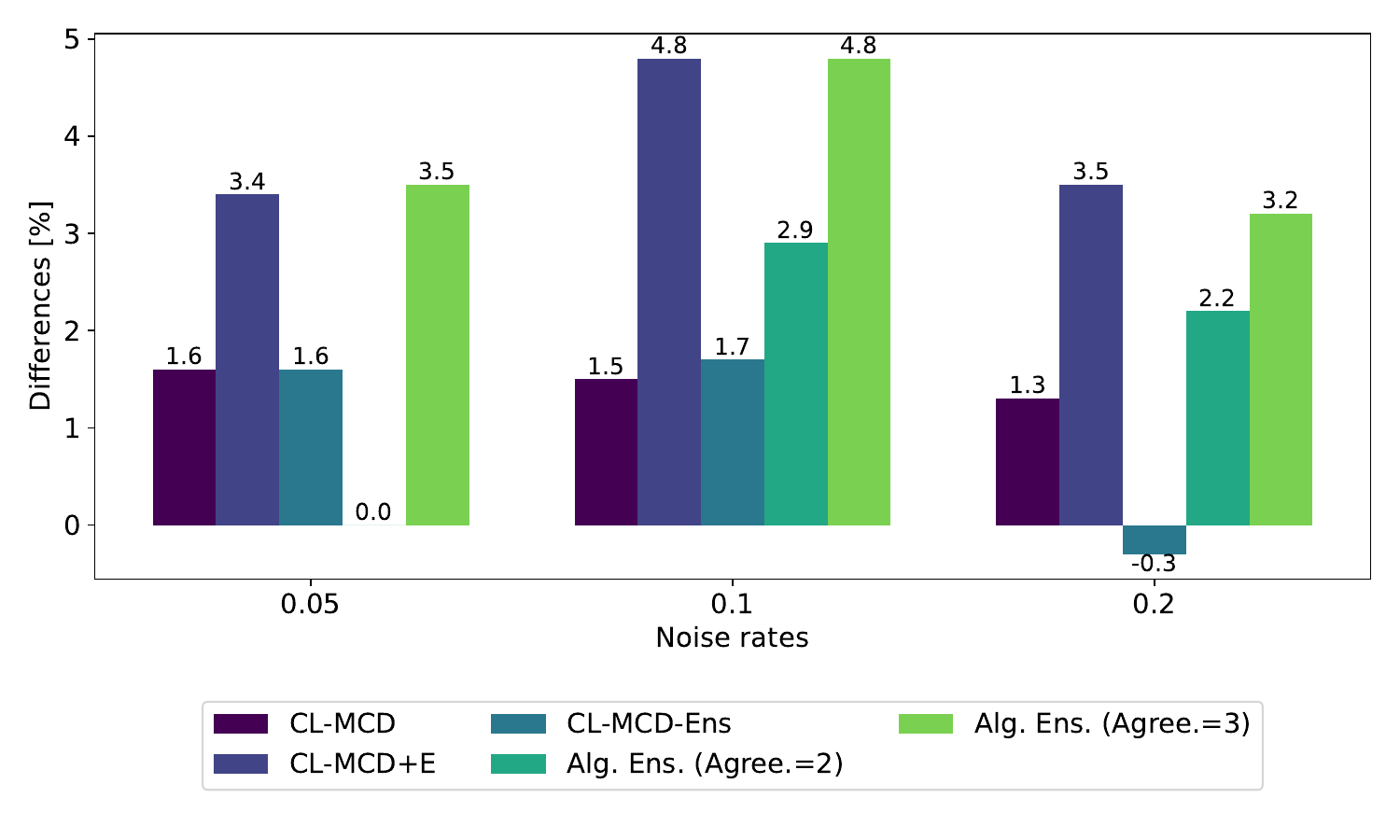}
    \caption{CIFAR-100}
     \end{subfigure}     
\begin{subfigure}[t]{0.48\textwidth}
         \centering
\includegraphics[width=\textwidth]{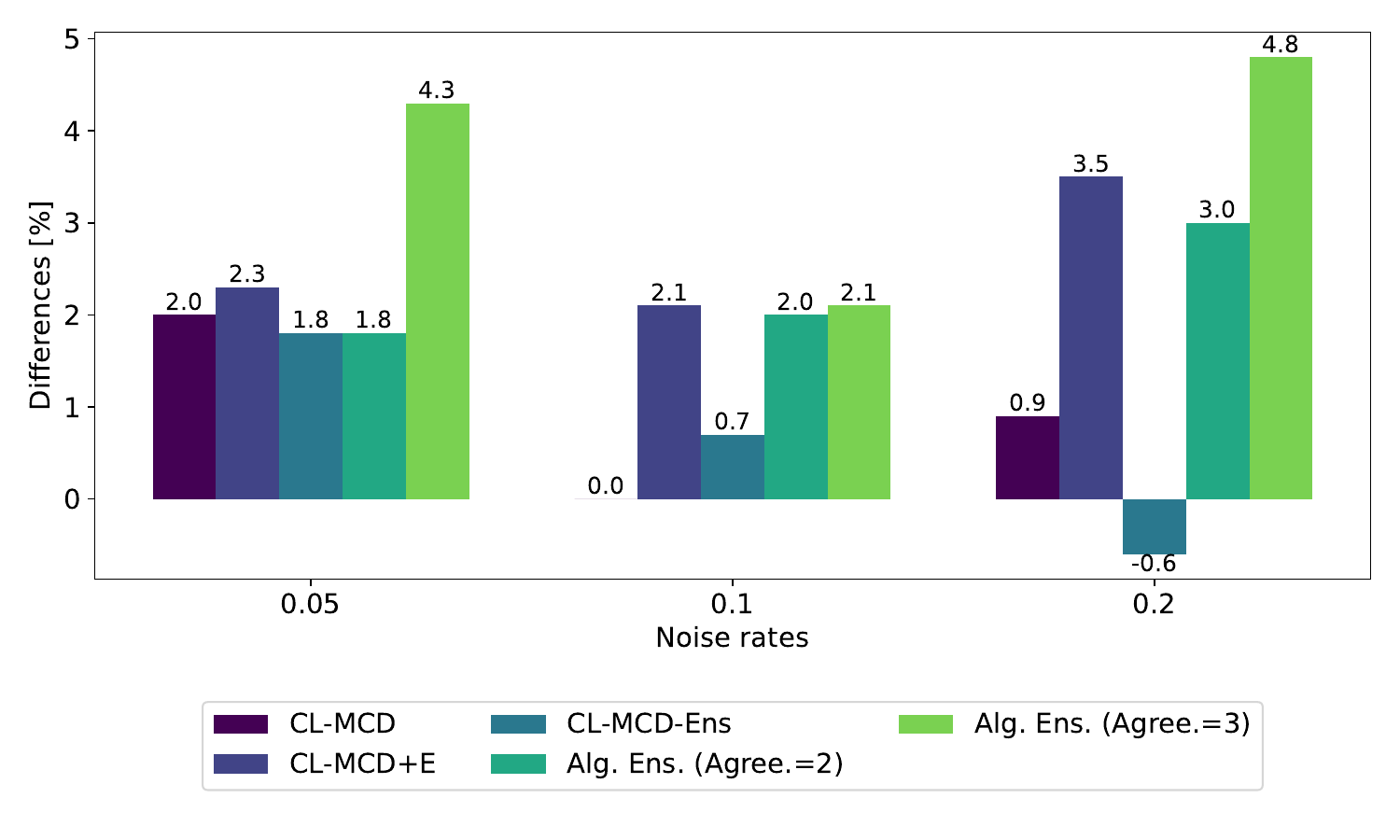}
    \caption{Tiny-ImageNet}
     \end{subfigure}     
     \caption[Comparing performance improvements of deep learning models based on uncertainty quantification-based label error detection and baselines]{Comparing performance improvements based on the proposed algorithms compared to CL-PBNR when excluding the label errors identified in Stage~1 in the training datasets. Test datasets remain unchanged for evaluation. Read: At noise rate $\tau_1=0.05$, cleaning with the CL-MCD algorithm leads to an accuracy gain of 0.4\% compared to cleaning with CL-PBNR.}
     \label{fig:stage-2-relative}
\end{figure*}

Importantly, the proposed algorithms again consistently outperform the state-of-the-art CL-PBNR algorithm in terms of the resulting model performance after detecting and removing label errors with the algorithms. 
In our experiments, removing label errors based on the proposed algorithms improved the mean performance (i.e., averaged over all considered noise rates) over the CL-PBNR algorithm by up to 0.3pp for MNIST, 0.9pp for CIFAR-10, and 2.6pp for CIFAR-100. For Tiny-Imagenet, excluding label errors did not improve the performance for any algorithm---however, our proposed algorithms still outperformed CL-PBNR by up to 2.0pp. 
Thus, overall, the proposed algorithms are not only more accurate in identifying label errors (see Evaluation Stage~1), but these improvements also translate to an increased performance of ML models based on excluding label errors.
We present the relative improvements of the developed algorithms in Figure~\ref{fig:stage-2-relative}.

Besides the performance improvements of the proposed algorithm, our experiments demonstrate that these algorithms exclude significantly less data as potential label errors than CL-PBNR. Together with the strong increases in precision scores we observed in the first stage of the evaluation, we can infer that the performance improvements compared to CL-PBNR can be attributed to a significant reduction of false positives. 
For example, at $\tau_3=0.2$, the CL-MCD-E algorithm achieves a 2.1\% better accuracy after cleaning than the baseline and removes 26.0\% fewer samples on the CIFAR-10 dataset. 
On the CIFAR-100 dataset, the Algorithm Ensemble (Agreement=3) outperforms the baseline by 3.5\% while removing 47.0\% fewer samples at $\tau_1=0.05$. 
For Tiny-ImageNet, at $\tau_1=0.05$, the Algorithm Ensemble (Agreement=3) achieves a 4.3\% increase in accuracy over the baseline while removing 41.8\% fewer samples.

\subsection{Sensitivity Analyses} 
\label{sec:general-results}

Throughout the following sensitivity analyses, we will investigate three important relations in label error detection. The first sensitivity analysis studies the relationship between the relative share of label errors in the dataset and the accuracy of label error detection. We hypothesize that the relative share of label errors in the dataset may influence the accuracy of detecting label errors since the distributions of correctly labeled and mislabeled data change, which might influence confident learning in general and UQLED in specific.
The second relationship is represented by the initial accuracy of the deep learning model and the accuracy in detecting label errors. We expect that a higher initial model accuracy translates to increased accuracy in label error detection as the model introduces fewer classification errors.
Finally, the third relationship deals with the influence of the accuracy in label error detection on the final model performance trained on the cleaned datasets. In line with previous work, we expect higher performance in label error detection to translate to higher final model performances based on high-quality input data.

\begin{figure}[h]
\centering
  \begin{minipage}[c]{0.6\textwidth}
    \centering
      \centering
     \begin{tabular}{l|ccc}
\toprule
Noise Rate $\tau$ & 0.05       & 0.1        & 0.2        \\ \midrule
MNIST             & 95.3\%     & 95.1\%     & 94.3\%     \\
CIFAR-10          & 59.4\%     & 63.7\%     & 64.3\%     \\
CIFAR-100         & 28.9\%     & 43.1\%     & 56.7\%     \\
Tiny-ImageNet     & 25.5\%     & 40.7\%     & 58.5\%     \\ \bottomrule
\end{tabular}%
  \end{minipage}
  \begin{minipage}[c]{0.35\textwidth}
    \centering
\includegraphics[width=0.8\textwidth]{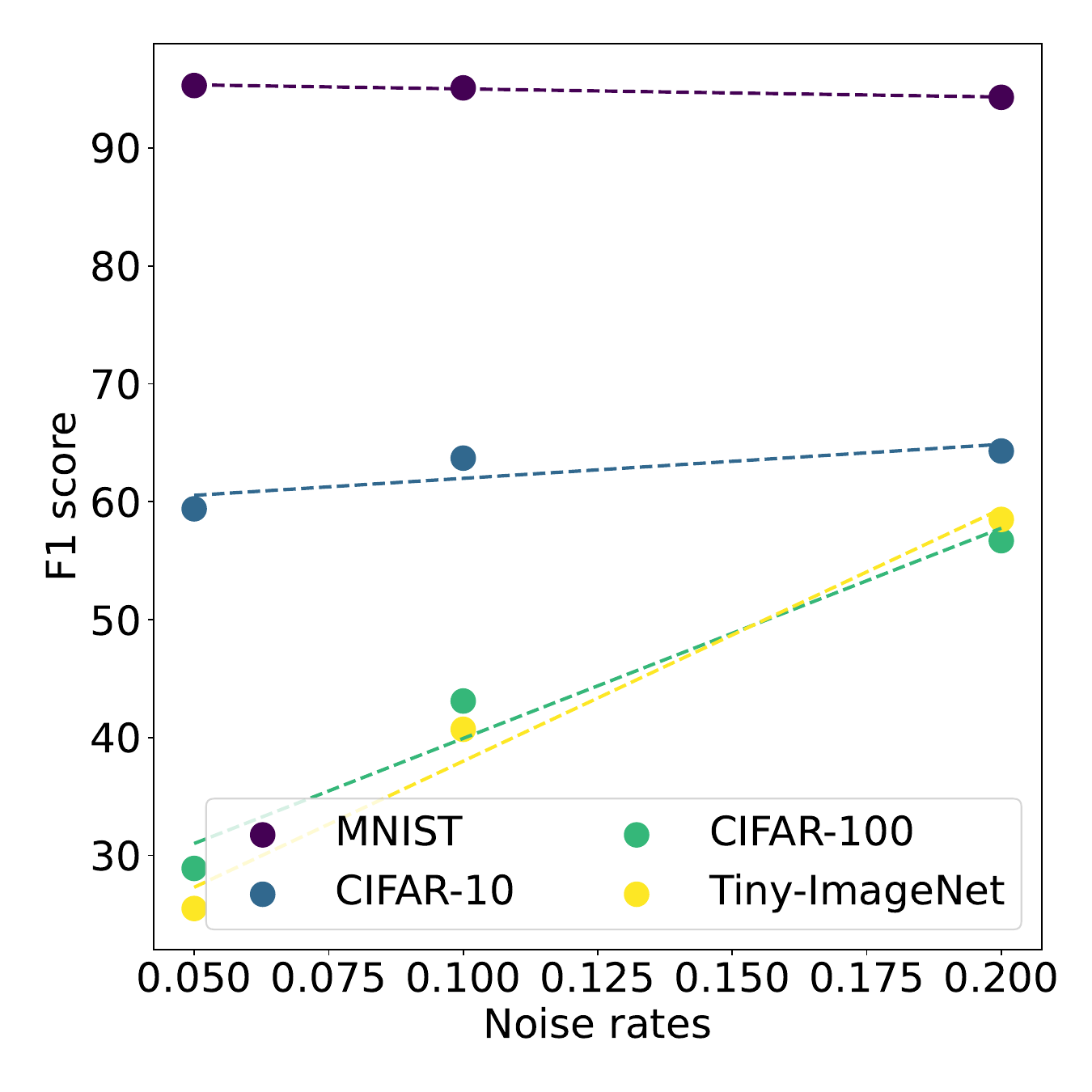}
  \end{minipage}%
  \caption[Mean F1 score across all algorithms per noise rate and dataset]{Mean F1 score across all algorithms per noise rate and dataset.}
\label{fig:average-f1-per-noise-rate-and-dataset}
 \end{figure}

\paragraph{Cross-Method Influence of Varying Noise Rates on Label Error Detection.}\quad
When comparing the performances across different label noise rates in Figure~\ref{fig:average-f1-per-noise-rate-and-dataset}, we observe a trend that for lower noise rates (i.e., smaller shares of label errors in the datasets), detecting such mislabeled data becomes more difficult. We observe this trend for CIFAR-10, CIFAR-100, and Tiny-Imagenet. For MNIST, the label error detection performance plateaued and appears less sensitive to variations in the noise rate. We explain this behavior with a generally high initial model performance on MNIST due to the simplicity of the dataset. 
In contrast, for datasets where the initial model performance is lower (e.g., CIFAR-100 and Tiny-ImageNet), the trend is particularly pronounced. 
For the CIFAR-100 dataset, the average F1 score reduces almost by half from 56.7\% at $\tau_3=0.2$ to 28.9\% at $\tau_1=0.05$ . Similarly, for Tiny-ImageNet, the average F1 score declines from 58.5\% at $\tau_3=0.2$ to 25.5\% at $\tau_1=0.05$.

Overall, the findings from this first sensitivity analysis align with previous work \cite{northcutt_pervasive_2021} and demonstrate that label error detection becomes more difficult for lower noise rates (i.e. when the relative share of mislabeled data is low). Importantly, our proposed algorithms have demonstrated particularly high performances in such settings (see Evaluation Stage~1) and are tailored to remedy such challenges.

\begin{figure}[h]
\centering
  \begin{minipage}[c]{0.6\textwidth}
    \centering
      \centering
      \footnotesize
   \begin{tabular}{lcc}
\toprule
& Initial Acc.   & F1 Score \\
\midrule
MNIST & 98.8\%  & 94.9\% \\
CIFAR-10 & 83.4\% & 62.5\% \\
CIFAR-100 & 64.8\%& 42.9\%  \\
Tiny-ImageNet  & 59.9\%  & 41.6\%\\
\bottomrule
\multicolumn{3}{p{\textwidth}}{Pearson correlation between the two set of scores is $r=0.98$. 
Two-sided t-tests with $n=4$, $df=2$ result in a test statistic of $t=6.37$ and a p-value of 0.02.} 
\end{tabular}
\end{minipage}
  \begin{minipage}[c]{0.35\textwidth}
    \centering
\includegraphics[width=0.8\textwidth]{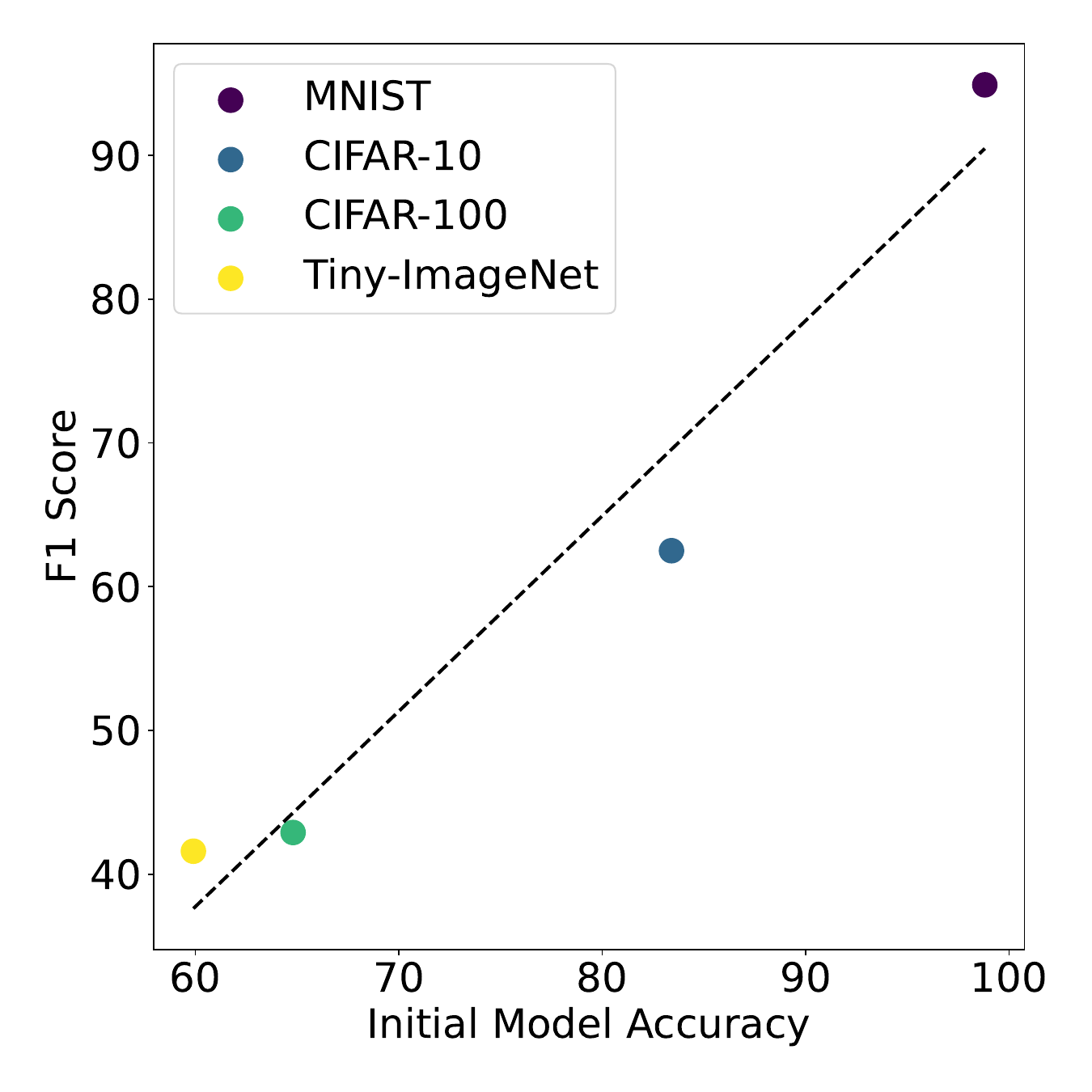}
  \end{minipage}%
\caption[Mean initial accuracy of the models and the mean F1 score across all algorithms and noise rates]{Mean initial accuracy of the models and the mean F1 score across all algorithms and noise rates for each dataset together with the correlation and significance measures.}
\label{fig:mean-acc-mean-f1}
 \end{figure}

\paragraph{Relationship Between Initial Model Accuracy and Label Error Detection Performance.}\quad
As a second sensitivity analysis, we investigate the relationship between the initial performance of the ML model and the accuracy of the label error detection. 
Recall that the initial model accuracy refers to the performance on the original dataset before identifying or removing any potential label errors. 
In Figure~\ref{fig:mean-acc-mean-f1}, we observe a strong correlation between the initial performance of the ML model and the label error detection performance across the datasets. 

We measure this correlation with the Pearson correlation coefficient $r$ \cite{schober_correlation_2018}, which ranges between -1 and 1, where a perfect correlation is represented by $r=1$. 
A correlation coefficient of zero indicates the absence of a linear relation between two variables. 
For the investigated relationship of the initial performance of the ML model and the label error detection performance, we observe a Pearson correlation coefficient of $r=0.98$, indicating a strong positive correlation. 
Note that, because $r$ is just an estimate of the population correlation $\rho$, it is necessary to test for significance \cite{turney_pearson_2022}.
For this purpose, we formulate the following hypotheses. The null hypothesis ($H_0$) refers to that the initial model accuracy and the label error detection performance are not significantly linearly correlated ($\rho=0$). In contrast, in the alternative hypothesis ($H_a$), both variables are significantly linearly correlated ($\rho \neq 0$).
The sample size $n$ corresponds to the number of datasets, i.e., $n=4$, and the degrees of freedom equals $df=n-2=2$. 
We leverage the two-tailed t-test to assess the significance of the correlation coefficient at a significance level of $\alpha=0.05$. 
We first calculate the test statistic (also called $t$-value) $t=\frac{r\sqrt{n-2}}{\sqrt{1-r^2}}$ \cite[p. 76]{cohen_statistical_1988}. 
We then calculate the $p$-value, which indicates how likely it is that the correlation between the two variables is random, using a two-sided $t$-distribution with $df=2$ \cite{turney_pearson_2022}. 
The test returns a p-value of 0.02, which means we can reject the null hypothesis ($H_0$) and accept the alternative hypothesis ($H_a$). 
This means that the correlation between the initial accuracy of the ML model and the label error detection performance is statistically significant. 
In other words, a higher initial accuracy of the ML model leads to a higher label error detection performance.

\begin{figure}[h]
\centering
  \begin{minipage}[c]{0.6\textwidth}
    \centering
      \centering
      \footnotesize
\begin{tabular}{lcc}
\toprule
                            & {F1 Score} & Final Acc.      \\ \midrule
CL-PBNR                     & {56.6\%}   & 77.3\%     \\
CL-MCD                      & {60.3\%}   & 77.8\%                 \\
CL-MCD-E                 & {63.3\%}   & 78.7\%            \\
CL-MCD-Ens             & {58.3\%}   & 77.7\%          \\
Alg. ens. (agree. = 2) & {61.2\%}   & 78.1\%          \\
Alg. ens. (agree. = 3) & {63.1\%}   & 78.6\%      \\ \bottomrule
\multicolumn{3}{p{\textwidth}}{Pearson correlation between the two set of scores is $r=0.98$. 
Two-sided t-tests with $n=6$, $df=4$ result in a test statistic of $t=10.03$ and a p-value of 0.001.} 
\end{tabular}
\end{minipage}
  \begin{minipage}[c]{0.35\textwidth}
    \centering
\includegraphics[width=0.8\textwidth]{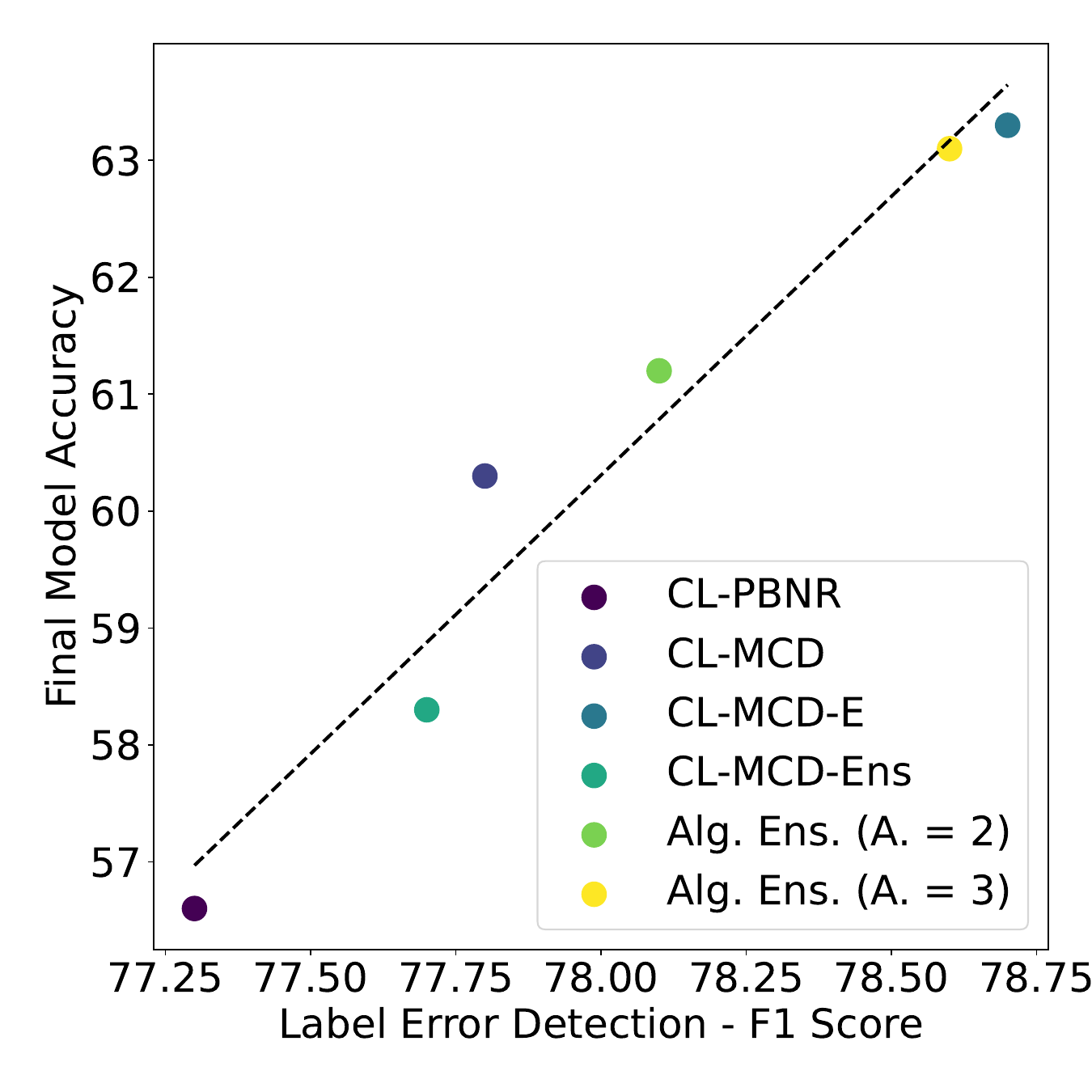}
  \end{minipage}%
\caption[Correlation between the mean label error detection performance and the mean final accuracy]{Correlation between the mean label error detection performance and the mean final accuracy after cleaning all training datasets with the algorithms.}
\label{fig:mean-f1-mean-final-acc}
 \end{figure}

\paragraph{Relationship between Label Error Detection Performance and Final Model Accuracy.}\quad
In the third sensitivity analysis, we focus on the relationship between the performance of the label error detection and the final model accuracy.
Recall that the final model accuracy refers to the performance of the ML model, which was trained on the training dataset after removing detected label errors. 
In Figure~\ref{fig:mean-f1-mean-final-acc}, we report the average label error detection performance (F1 score) for each algorithm together with the average final accuracy averaged across all datasets. 
We observe that algorithms that perform better in detecting label errors lead to an improved final model accuracy. 
We again measure the Pearson correlation and test the statistical significance. 

The calculated correlation coefficient for the given sample is $r=0.98$. This suggests that the label error detection performance and the final model accuracy are strongly correlated---i.e., the better the label error detection, the higher the final model accuracy. 
To test the significance of the correlation, we again employ a two-sided t-test at a significance level of $\alpha=0.05$. 
The sample size $n$ corresponds to the number of algorithms, i.e., $n=6$, and the degrees of freedom equals $df=n-2=4$. 
We define the null hypothesis ($H_0$) as the absence of a significant linear correlation between the label error detection performance and the final model accuracy ($\rho=0$). The alternative hypothesis ($H_a$) refers to the presence of a significant linear correlation ($\rho \neq 0$).
We obtain a p-value associated with the test statistic of $p=0.001<0.05=\alpha$, such that the null hypothesis $H_0$ can be rejected, and the alternative hypothesis $H_a$ can be accepted. 
Thus, the label error detection performance and the final model accuracy are significantly linearly correlated.
In other words, the higher the performance in label error detection, the higher the final model accuracy.

Overall, our sensitivity analyses reveal three important properties of the examined algorithms for label error detection. 
First, the higher the relative share of label errors in the dataset, the higher the detection accuracy is.
Second, the more accurate the models are on the original, uncleaned dataset, the higher the accuracy in the detection of label errors. 
Third, the higher the accuracy in label error detection is, the higher the model performance on the cleaned datasets. 
Our analyses underline the importance of removing label errors from datasets to achieve a significant increase in the final model accuracy.

\section{Discussion and Conclusion}\label{sec:conclusion}

The goal of this work was to improve state-of-the-art label error detection, i.e., {CL}, with the help of model uncertainty.  
For this purpose, different label error detection algorithms based on {MCD} inference were developed and experimentally evaluated against a baseline (CL-PBNR) within a two-step evaluation procedure. 
Using {MCD} allowed us to incorporate model uncertainty into our algorithms and thus account for it during the label error detection. 
In addition, ensemble learning techniques were used to further improve the performance and generalizability of our proposed algorithms.
After passing the preliminary evaluation on the test sets of four prominent {ML} benchmark datasets in Stage~{1}, the developed algorithms were then comprehensively evaluated in Stage~{2} on the entire benchmark datasets. 
Besides the label error detection performance, which we measured by the F1 score, we also assessed the practical implications in terms of (positive) effects on the model accuracy after the removal of the label errors identified. 
The results of these experiments will be interpreted and discussed in the following. 
Thereby, we will first examine the label error detection performance before focusing on the implications of our algorithms on the model accuracy.
The results obtained in Stage~{2} show that our proposed algorithms outperform the baseline (CL-PBNR) in terms of label error detection performance across all datasets and noise rates. 
The only exceptions, which we will come to later, are the CL-MCD-E algorithm and the Algorithm Ensemble (Agreement=3) at noise rate $\tau_3=0.2$ on the MNIST dataset.
The reason for the overall better label error detection performance, i.e., the overall better F1 score, can be attributed to the higher precision of all our algorithms across all datasets and noise rates. In the following, we highlight important implications resulting from our work and discuss its limitations, followed by the concluding remarks. 

\subsection{Replacing Softmax Probabilities with Monte Carlo Dropout Probabilities}

The overall higher precision of our algorithms can mainly be justified by the use of {MCD}. 
When replacing softmax probabilities (see CL-PBNR) with {MCD} probabilities (see CL-MCD), the average precision increases by 12.2\% across all noise rates and datasets. 
The reason for the increase in precision is that samples that are difficult to classify for the model (i.e., higher model uncertainty) are filtered out in the first step of {CL} (\textit{Count}). 
Thus, these samples cannot be counted as label errors in the confident joint. 
Consequently, by using {MCD} probabilities, fewer samples for which the model is uncertain are incorrectly flagged as label errors, which increases the precision of the algorithms compared to CL-PBNR.
In contrast to softmax probabilities, MCD probabilities additionally allow the employ of uncertainty measures, such as entropy, on top. 
This further increases the performance of the label error detection by, on average, 25.8\% in precision when comparing the CL-MCD-E algorithm with the baseline. 
Thus, MCD probabilities not only improve the performance in general, but they additionally facilitate the utilization of additional metrics to refine label error detection.

\subsection{Prerequisites for Label Error Detection}

An important ingredient for an improvement in the final model accuracy is successful label error detection.  
Our experiments have demonstrated that the performance in label error detection is correlated with the initial model accuracy. 
This means that when the initial accuracy of the model is low, as, for example, with the Tiny-ImageNet dataset, the absolute performance in detecting label errors is low as well---eventually preventing an improvement in the final model accuracy. 
The lower final model performance is reflected in the label error detection performance, where the precision is substantially lower compared to, e.g., the CIFAR-10 dataset. This causes many images to be incorrectly removed, and the final accuracy is negatively affected by removing label errors from the training dataset.
Thus, in order to achieve improvements in the final model accuracy, sufficient initial model accuracy is required.
Overall, this emphasizes the importance of the complementarity and co-existence of data-centric and model-centric AI, where model-centric AI provides a substantial initial model accuracy, which is required for data-centric detection and removal of label errors.
In addition, our experiments revealed that label error detection algorithms are vulnerable to low noise rates (i.e. when only a few label errors exist in the training data). In these cases, we observed the lowest performance of all algorithms. Since in real-world and benchmark datasets, the noise rates typically range between 1\% and 10\%, it is particularly important to account for this behavior. Thus, another prerequisite for successful label error detection is algorithms that reliably identify label errors even in very imbalanced settings. We understand UQ-LED algorithms as a first approach in this direction.

\subsection{Efficiency Considerations}

Our findings show that removing label errors from training datasets can significantly increase the performance of deep learning models. However, researchers and practitioners particularly need to monitor the number of data samples that are removed by the algorithms.  
Our proposed algorithms generally exhibit a higher precision, which means that fewer samples are removed from the dataset by mistake. 
If, however, the precision of a label error detection algorithm is low at a very high recall, too many images are incorrectly removed during cleaning. 
This can have an overall negative effect on model training since there might be too little training data per class available to learn the class representations correctly.
Consider the following example. 
We synthetically add 5,000 label errors to a dataset with 50,000 samples in total (i.e., noise rate of 10\%).
A label error detection algorithm with a recall of 100\% and a precision of 25\% will incorrectly remove 25,000 images, such that in the best case, 20,000 correctly labeled images are removed. 
Then, instead of 50,000 images, the model has only 30,000 correctly labeled images available for training, which might be insufficient to learn the class representations.
However, if the algorithm achieves a substantially lower recall of 70\% and a slightly higher precision of 35\%, the algorithm removes only 10,000 images. This means that in the best case, only 5,000 correctly labeled images are removed, and a resulting 45,000 correctly labeled images are available for model training. 
These considerations are not only important from a performance perspective but also from an efficiency perspective. 
Labeling large bodies of data for supervised machine learning is costly, especially in high-stakes domains where human experts have to annotate the data. 
Removing correctly labeled images from the datasets is associated with significant costs in terms of unnecessary labeling effort. 
This makes a higher precision in label error detection particularly relevant. 

\subsection{Limitations}
Like any work, this study is not free of limitations. The original {CL} method of \cite{northcutt_confident_2019}, which we used as the foundation for our own algorithms, is fully model agnostic and can be used not only for neural networks but also, for example, for regression problems, support vector machines, or decision trees. 
Our proposed algorithms build on Monte Carlo Dropout and, thus, are specific to {DNN}s. 
However, we want to emphasize that the algorithms are agnostic to different architectures of deep neural networks since dropout layers can be added to any {DNN}.
In addition, our results may be affected by actual label errors that have been inherently present in the original datasets before adding synthetically generated label noise. This may result in the following two effects on the performance measurements of label error detection performance. 
First, if label errors have been present before adding synthetic errors and the algorithms successfully detect them, their \textit{precision} would be negatively affected because, in contrast to the synthetically added label errors, the already existing label errors were not marked as such by us before the evaluation. 
As a result, the absolute precision of all algorithms may be higher than we measured. 
Notably, this applies to all algorithms, including the CL-PBNR baseline, thus not affecting the comparison of algorithms and our findings.
Second, pre-existing label errors could automatically be corrected when generating synthetic label errors. 
Since human annotators often mislabel classes that are similar, our label noise generator considers class similarities during noise generation. 
As a result, the label errors unintentionally corrected during the noise generation cannot be identified as such by the different algorithms. 
Consequently, in case of such an event, the \textit{recall} of the algorithms would be reduced since we have previously marked the now corrected samples as label errors.
Again, this limitation applies equally to all algorithms. 
Thus the overall core message of this work, that {MCD}-based algorithms with high precision are preferable for label error detection, remains. 
If future work would like to address this limitation, it would be a good idea to clean the dataset using one of the algorithms before adding label noise. 
However, we decided against this to avoid influencing our results with the choice of the cleaning algorithm.

\subsection{Concluding Remarks}
Supervised machine learning is heavily influenced by the quality of the labeled data. Correct labels are an essential prerequisite for accurate and reliable model output. 
Removing erroneous labels from the training dataset is a vital data-centric approach to improve the model performance, which we enhance by building on methods for uncertainty quantification based on approximations of Bayesian neural networks. 
Our findings suggest that uncertainty quantification-based label error detection can significantly increase the performance of machine learning models and, thus, underlines the importance of well-curated datasets for supervised machine learning.

\newpage


\vskip 0.2in
\bibliography{refs}
\bibliographystyle{theapa}

\end{document}